\documentclass[11pt]{article}

\usepackage{multirow}
\usepackage{booktabs}
\usepackage{multicol}
\usepackage{amsmath}
\usepackage{tcolorbox}
\usepackage{subfig}
\usepackage[final]{acl}
\usepackage{todonotes}
\usepackage{times}
\usepackage{latexsym}

\usepackage[T1]{fontenc}

\usepackage[utf8]{inputenc}
\usepackage{microtype}

\usepackage{inconsolata}
\usepackage{enumitem}
\usepackage{graphicx}

\title{Graph-Guided Passage Retrieval for Author-Centric Structured Feedback}

\author{
\textbf{Maitreya Prafulla Chitale}$^{1}$\thanks{Equal contribution.} \quad 
\textbf{Ketaki Mangesh Shetye}$^{1}$\footnotemark[1] \\
\textbf{Harshit Gupta}$^{2}$ \quad 
\textbf{Manav Chaudhary}$^{3}$ \quad 
\textbf{Manish Shrivastava}$^{1}$ \quad
\textbf{Vasudeva Varma}$^{1}$ \\
$^1$IIIT Hyderabad \quad
$^2$Microsoft, India \quad
$^3$Sentisum \\
\texttt{\{maitreya.chitale, ketaki.shetye\}@research.iiit.ac.in} \\
\texttt{guptaharshi@microsoft.com} \\
\texttt{manav@sentisum.com} \\
\texttt{\{m.shrivastava, vv\}@iiit.ac.in}
}

\begin{document}
\maketitle
\begin{abstract}
Obtaining high-quality, pre-submission feedback is a critical bottleneck in the academic publication lifecycle for researchers. We introduce AutoRev, an automated author-centric feedback system that generates structured, actionable guidance prior to formal peer review. AutoRev employs a graph-based retrieval-augmented generation framework that models each paper as a hierarchical document graph, integrating textual and structural representations to retrieve salient content efficiently. By leveraging graph-based passage retrieval, AutoRev substantially reduces LLM input context length, leading to higher-quality feedback generation. Experimental results demonstrate that AutoRev significantly outperforms baselines across multiple automatic evaluation metrics, while achieving strong performance in human evaluations. Code will be released upon acceptance.
\end{abstract}

\section{Introduction}
\label{introduction}
Recent advancements have focused on assisting reviewers and directly generating peer reviews \cite{yu-etal-2024-automated, aitymbetov-zorbas-2025-autonomous, darcy2024margmultiagentreviewgeneration, idahl-ahmadi-2025-openreviewer}. Concurrently, there has been growing interest in examining the strengths and limitations of such systems, particularly the use of large language models (LLMs) in this domain \cite{ye2024yetrevealingrisksutilizing}. Relying on LLMs to generate or inform final review decisions remains risky due to issues of bias, overconfidence, and superficial feedback and these decision-oriented tools lack utility for authors in the pre-submission phase. Consequently, we advocate for a new paradigm towards author-centric support. We introduce AutoRev, an author-centric system that generates structured, actionable pre-submission feedback containing \textit{Summary}, \textit{Strengths}, \textit{Weaknesses}, and \textit{Questions}, without producing acceptance recommendations or replacing human peer review.

\begin{figure}[tbp]
  \centering
  \includegraphics[width=0.475\textwidth]{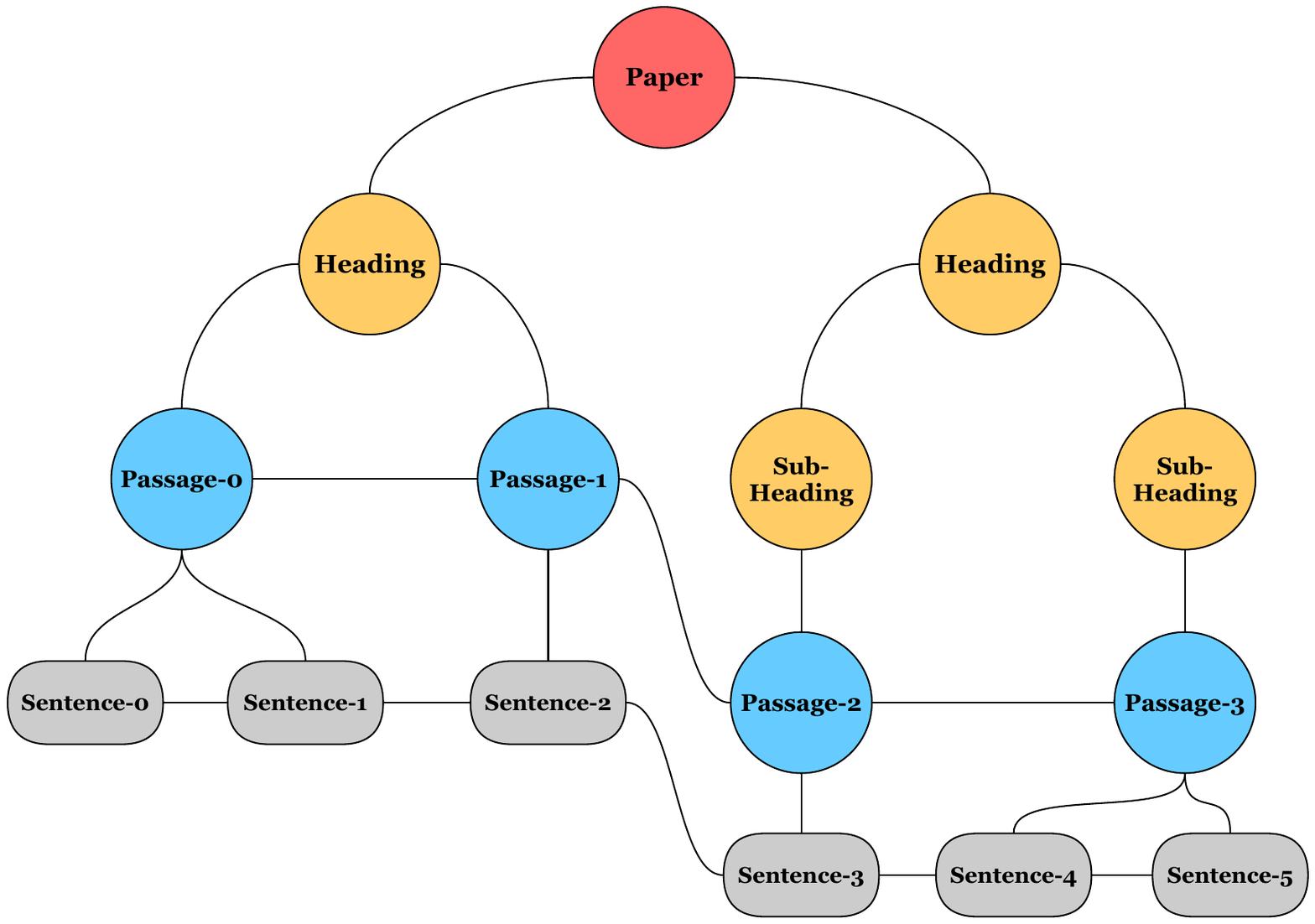}
  \caption{Example document graph with nodes for headings, sub-headings, passages, and sentences. This structure allows AutoRev to capture hierarchical and sequential dependencies across sections.}
  \label{fig:toy-example}
\end{figure}

With the rise of Graph Neural Networks, the use of graphs as \textit{extractors} or as \textit{guides} for facilitating downstream tasks has recently gained traction in the NLP community \cite{chitale-etal-2025-discograms, gao-etal-2025-mixed, agarwal-etal-2025-hybrid, zhu2025graphbasedapproachesfunctionalitiesretrievalaugmented}. Inspired by this, we leverage the graph’s capability as an extractor of passages to significantly reduce the input token length to the LLM. Reducing the token length of the input is shown to enhance LLM performance \cite{hsieh2024ruler}.
To summarize our main contributions:
\begin{enumerate}
    \item We propose an author-centric system for generating concise, structured pre-submission feedback for academic papers.
    \item We introduce a graph-based retrieval framework that learns to identify salient passages, reducing LLM input length while improving feedback quality.
    \item We validate our approach using comprehensive human evaluations to assess feedback quality and factual grounding, in addition to standard automatic metrics.
\end{enumerate}

\section{Leveraging Peer Reviews as Proxy for Pre-Submission Feedback}
\label{peer-reviews-as-proxy}
In the absence of large-scale, publicly available datasets explicitly designed for pre-submission feedback, we use human-written peer reviews as a proxy, as they provide expert critiques of writing, methodology, and presentation aligned with the guidance authors seek prior to submission. In particular, the \textit{Weaknesses} and \textit{Questions} sections offer actionable critiques and clarifications. Since the dataset spans papers at varying stages of writing maturity, the associated reviews reflect feedback targeted at authors with differing levels of experience. We further support this choice through an impact-oriented user study (Section~\ref{sec:impact}).
\section{Related Works}
Recent work has explored the use of NLP techniques to support scholarly writing and review workflows, including reviewer selection \cite{aitymbetov-zorbas-2025-autonomous}, meta-review generation \cite{10.1007/s00799-023-00359-0}, and automated review generation \cite{yu-etal-2024-automated}. Many of these approaches rely on large language models, ranging from modular agent-based systems \cite{darcy2024margmultiagentreviewgeneration} to fine-tuned review generators \cite{idahl-ahmadi-2025-openreviewer}. In parallel, a separate line of work focuses on author-assistive tools for specific writing sub-tasks, such as limitation suggestion \cite{10.1007/978-3-031-70344-7_7}, citation recommendation and literature discovery \cite{gu-hahnloser-2023-scilit}, and drafting individual sections like related work \cite{hu-wan-2014-automatic} or introductions \cite{liebling-etal-2025-towards}. While these are effective within their scoped objectives, they are either reviewer-facing or narrowly targeted, and often operate on full-paper inputs at inference time. As a result, they raise concerns around hallucination, superficial feedback, computational cost, and overconfidence in model-generated judgments \cite{ye2024yetrevealingrisksutilizing}, while failing to provide holistic, structured feedback that supports authors prior to submission.

\citet{yu-etal-2024-automated} introduce an automated paper reviewing framework (SEA) that standardizes reviews using GPT-4 \cite{openai2024gpt4technicalreport} consolidation and proposes a self-correction strategy during fine-tuning. We adopt SEA as our baseline as it is trained on the publicly available ICLR 2024 dataset and generates the same structured fields as AutoRev, among others.




Studies have explored graph-based extensions to Retrieval-Augmented Generation (RAG) \cite{10.5555/3495724.3496517} to improve grounding and reasoning in LLMs. \citet{zhu2025graphbasedapproachesfunctionalitiesretrievalaugmented} provide a comprehensive taxonomy of Graph-RAG systems based on database construction, retrieval mechanisms, and processing pipelines. Within this framework, AutoRev represents a domain-specific Graph-RAG approach that constructs a hierarchical graph directly from a scientific paper and employs a learning-based Graph Attention Network \cite{veličković2018graph} for passage retrieval.


\section{Dataset}
\label{sec:data}
The dataset used in our study is borrowed from \citet{yu-etal-2024-automated}, comprising papers submitted to various conferences along with multiple peer reviews per paper. Among these, the ICLR 2024 and NeurIPS 2023 datasets are publicly released, though the GPT-4-consolidated reviews for NeurIPS 2023 are not available. Hence, our analysis focuses only on the ICLR 2024 papers. Table \ref{tab:dataset} summarizes statistics for the ICLR 2024 papers \cite{yu-etal-2024-automated}.

\begin{table}[htbp]
    \begin{tabular}{lc}
        \toprule
        \textbf{Attribute} & \textbf{Value} \\
        \midrule
        Total Papers & 5,653 \\
        Average Tokens per Paper & 9,815 \\
        Total Reviews & 21,839 \\
        Acceptance Rate & 37\% \\
        Domain & Machine Learning \\
        \bottomrule
    \end{tabular}
\caption{Dataset statistics for the ICLR 2024 dataset, as reported in \citet{yu-etal-2024-automated}.}
\label{tab:dataset}
\end{table}

\noindent Each paper is associated with multiple reviews containing sections such as Summary, Strengths, Weaknesses, Questions, Soundness, Presentation, Contribution, Confidence, and Rating. We retain only the Summary, Strengths, Weaknesses, and Questions sections, as these encode constructive feedback for authors-centric pre-submission support (see Section~\ref{peer-reviews-as-proxy}), and exclude rating-based fields that correspond to final decision-making. For training, we use the GPT-4 consolidated reviews released by \citet{yu-etal-2024-automated}, which merge multiple human-written reviews into a single structured review per paper.

To assess cross-venue performance, we curate a supplementary dataset of 10 papers from COLM and NeurIPS 2025 (CNT\_10). This set complements the ICLR 2024 test data (ICT) for the human evaluation detailed in Section~\ref{human-evaluation}.
\begin{figure*}[htbp]
    \centering
    \includegraphics[width=0.9\textwidth]{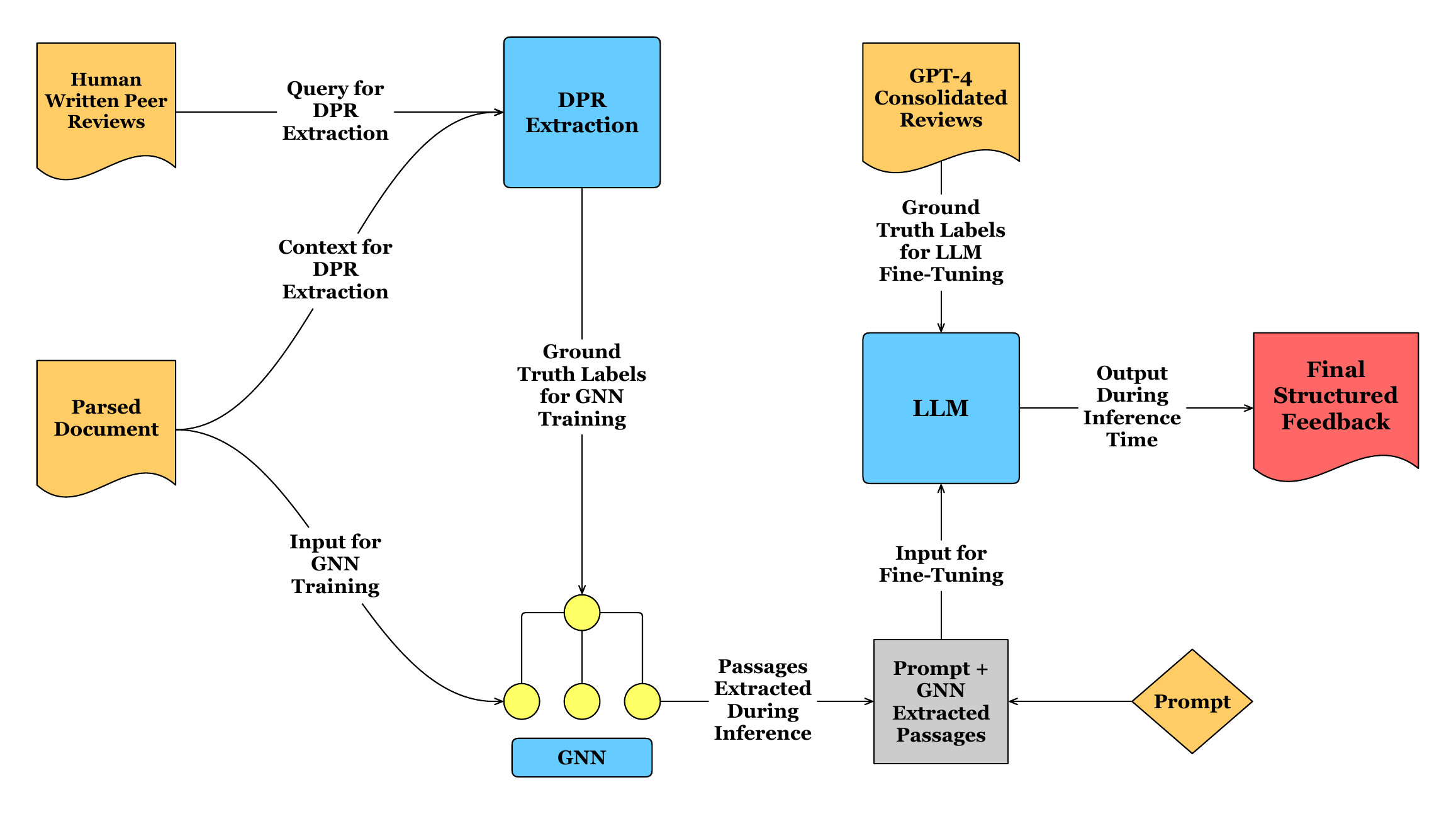}
    \caption{The system begins with a parsed document from which DPR extracts passages for GNN training. The GNN learns to identify key passages relevant to structured feedback generation. The trained GNN selects important passages without requiring human-generated feedback. These passages, combined with a structured prompt, are input to fine-tune an LLM. The fine-tuned LLM generates the final structured feedback, without external grounding.}
    \label{fig:main-methodology}
\end{figure*}
\section{Methodology}
In this section, we describe the implementation of our proposed framework. Section~\ref{dpr-passsage_extraction} describes the initial passage extraction using Dense Passage Retrieval (DPR) \citep{karpukhin-etal-2020-dense}. The extracted passages are used as training data in Section~\ref{gnn-training}, to train a Graph Neural Network (GNN) that learns to extract relevant passages from a research paper. The trained GNN enables our model to perform competitively on the test set without the need for grounding on human-generated feedback, as discussed in Section~\ref{grounding section}. Post that, we describe our process of fine-tuning LLMs to generate the final structured feedback, in Section~\ref{llm-finetuning}. An overview of our methodology can be found in Figure~\ref{fig:main-methodology}.
\subsection{DPR-based Passage Extraction}
\label{dpr-passsage_extraction}
Dense Passage Retrieval (DPR) is a neural retrieval model trained primarily on question-answering datasets that encodes queries and passages into dense vector representations. Given an input query, DPR retrieves relevant passages based on semantic similarity in the embedding space. We use a pre-trained DPR model to identify passages corresponding to the \textit{Summary}, \textit{Strengths}, \textit{Weaknesses}, and \textit{Questions} sections of reviews for each paper in the training set. Specifically, we use the context encoder\footnote{\url{https://huggingface.co/facebook/dpr-ctx_encoder-multiset-base}} and query encoder\footnote{\url{https://huggingface.co/facebook/dpr-question_encoder-multiset-base}} from the \textit{multiset-base} DPR variant.


First, we segment each paper into discrete passages, where each passage is defined as a text segment split at newline characters (\texttt{\textbackslash n}). Each of these passages is treated as a potential context candidate for DPR. Specifically, for each section of the feedback, we concatenate $k$ consecutive sentences (with $k = 1$, $3$, or $5$) to form a single query, to capture a broader contextual meaning. For each query, the top-$m$ most relevant passages (with $m = 3$ or $5$) are retrieved using DPR. We then consolidate all unique passages retrieved across these multiple queries for each paper.

This final set of passages represents the most relevant excerpts from the full paper that are semantically linked to the feedback content. These extracted passages serve as the ground truth labels for training the GNN, as described in Section~\ref{gnn-training}
\subsection{Graph Construction for Representing a Paper}
\label{graph-construction}
The idea of graph construction employed in our work is inspired by \citet{zhao-etal-2024-hierarchical}. We borrow their idea of representing a document as a hierarchical graph but extend it beyond sentence level. The graph is constructed directly from the parsed paper and is designed to generalize across diverse academic writing styles and section hierarchies.

Formally, each paper is represented as an undirected graph $\mathcal{G} = (\mathcal{V}, \mathcal{E})$, where nodes correspond to textual units at different granularities, including the paper root, section headings, sub-headings, passages, and sentences. Edges encode two complementary relationships: (i) hierarchical links ($\mathcal{E}_{hier}$) that reflect the document’s structural nesting and (ii) sequential links ($\mathcal{E}_{seq}$) that preserve the linear ordering of passages and sentences. Together, these connections allow information to propagate both vertically across document structure and laterally across adjacent text. A schematic illustration of the document graph is shown in Figure \ref{fig:toy-example}. Full details of node types, edge definitions, and construction procedures are provided in Appendix \ref{app:graph-constr}.

This design explicitly mitigates over-squashing in tree-like graphs, where exponential neighborhood growth leads to information bottlenecks at the root \citep{alon2021on, topping2022understanding}. Following the rewiring strategies of \citet{topping2022understanding}, we augment the hierarchical backbone with sequential edges to facilitate lateral information flow and reduce reliance on hierarchical connections. Additionally, the inherent structure of scientific papers results in graphs of shallow depth (typically $\leq 5$ hops), limiting long-range dependency compression and preserving fine-grained information from leaf nodes.


\subsection{GNN Training for Passage Retrieval}
\label{gnn-training}
To train a GNN on the graph constructed in Section~\ref{graph-construction}, we take the data generated in Section~\ref{dpr-passsage_extraction}, which maps each paper $p_i$ to a set of important passages, \( \mathcal{O}_i = \{ o_1, o_2, \dots, o_n \} \). Each passage is important for generating a feedback for the paper it is associated with. This mapping is used as a training set for training the GNN. The node embeddings for sentence nodes are initialized with embeddings from a Sentence-Transformer \cite{reimers-gurevych-2019-sentence} model, namely \textit{all-mpnet-base-v2}. The node embeddings for all other nodes are initialized with zero embeddings of dimension \( 768 \), matching that of the Sentence-Transformer model.

We train a Graph Attention Network (GAT) \cite{veličković2018graph} over the document graph to classify passage nodes as feedback-relevant or not.
The GNN is trained using various configurations based on the values of \(k\) and \(m\), as described in the DPR process in Section~\ref{dpr-passsage_extraction}. These trained GNN models are denoted as GNN \((k, m)\).
Additional implementation details related to this section are available in Appendix~\ref{sec:appendix-gnn}.

\begin{figure}[htbp]
  \centering
  \includegraphics[width=0.475\textwidth]{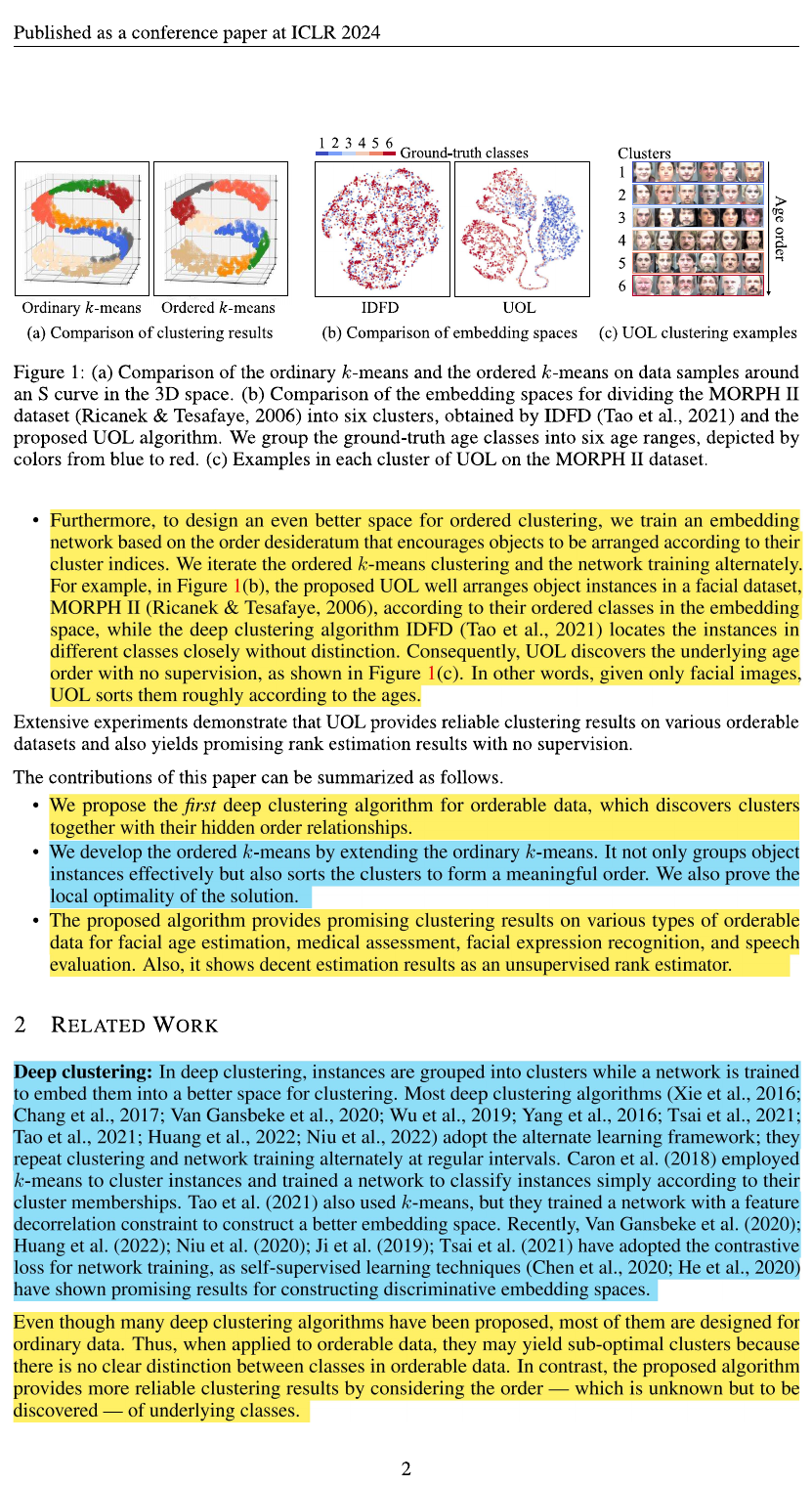}
  \caption{Excerpt of a test sample where passages highlighted in cyan colour refer to relevant passages identified by GNN only and yellow colour refer to relevant passages identified by both GNN and DPR}
  \label{fig:DPRvsGNN}
\end{figure}

As illustrated in the example from our test set (Figure~\ref{fig:DPRvsGNN}), the DPR-based approach fails to retrieve a key passage related to the main contributions of the paper, whereas our GNN successfully identifies it. Therefore, graph-based representation, which leverages both sequential and hierarchical node connections, effectively identifies the salient passages required to enhance the feedback generation process. The GNN extracted passages, along with a structured prompt, serve as the input in the LLM fine-tuning stage, as discussed in Section~\ref{llm-finetuning}.
\subsection{Grounding and Inference Efficiency}
\label{grounding section}
For practical deployment, our GNN-based method offers a crucial advantage by operating autonomously at inference time. Unlike DPR-based approaches, which rely on human-generated feedback to act as queries, that may be unavailable in real-world scenarios, our system identifies salient passages directly from the document graph, thereby eliminating the dependency on external grounding.

Unlike SEA baselines that process the full text, AutoRev generates feedback from a reduced set of GNN-selected passages. This approach significantly lowers computational cost and input length. As noted by \citet{hsieh2024ruler}, shorter contexts enhance LLM performance, which is also reflected in our higher scores compared to baselines (Section~\ref{resultsandanalysis}). Table~\ref{tab:gnn_reduction_stats} details the specific reductions in tokens and passages.
\begin{table}[htbp]
    \centering
    \begin{tabular}{lcc}
    \toprule
    \textbf{Method} & \textbf{Avg Token Len} & \textbf{Avg Passages} \\
    \midrule
        Full Paper & 9445.92 & 47.07 \\
        GNN \((5, 5)\)     & 4147.15 & 29.49 \\
        GNN \((5, 3)\)     & 2899.59 & 21.00 \\
    \bottomrule
\end{tabular}
\caption{Average token length (Avg Token Len) and Average passage count (Avg. Passages) for a full paper, our best-configuration: GNN \((5, 5)\), and our fastest-configuration: GNN \((5, 3)\).}
\label{tab:gnn_reduction_stats}
\end{table}
\subsection{Fine-Tuning Large Language Models for Structured Feedback Generation}
\label{llm-finetuning}
General-purpose LLMs are not trained specifically to generate structured feedback for academic research papers. Our motivation for fine-tuning is to ensure that the models learn to generate qualitatively sound, factual and structured feedback.

We select \textit{Mistral-Instruct-v0.2} (Mistral) \cite{jiang2023mistral7b} to align with our SEA-EA baseline, and state-of-the-art (SOTA), open-weight model, \textit{Llama 3.1 8B} (Llama) \cite{grattafiori2024llama3herdmodels} for its strong text generation capabilities. To efficiently adapt these models for structured feedback generation while ensuring computational feasibility, we employ Low-Rank Adaptation (LoRA) \cite{hu2022lora} for finetuning.

We use the trained GNN, as described in Section~\ref{gnn-training}, to retrieve relevant passages from each paper, which are provided as input to the LLM during fine-tuning. For supervision, we use the \textit{Summary}, \textit{Strengths}, \textit{Weaknesses}, and \textit{Questions} sections of human-written peer reviews for ICLR 2024 papers, consolidated using GPT-4 and released as part of the dataset introduced by \citet{yu-etal-2024-automated}. These consolidated reviews serve as structured feedback targets for training the LLM. The dataset split is summarized in Table~\ref{tab:dataset-split}.
\begin{table}[htbp]
    \centering
    \begin{tabular}{lcc}
    \toprule
    \textbf{Dataset Split} & \textbf{Number of Samples} \\
    \midrule
    Training Set & 1,765 \\
    Test Set     & 271 \\
    \bottomrule
    \end{tabular}
\caption{Number of samples used for training and testing sourced from \citet{yu-etal-2024-automated}}
\label{tab:dataset-split}
\end{table}
Both models are fine-tuned on passages retrieved via varying GNN \((k, m)\) configurations (Section~\ref{gnn-training}), denoted as Mistral \((k, m)\) and Llama \((k, m)\). Prompts and hyperparameters are detailed in Appendix~\ref{sec:appendix-ft}.
\section{Results and Analysis}
\label{resultsandanalysis}
\begin{table*}[htbp]
    \centering
    \begin{tabular}{l ccc ccc c c}
        \toprule
        \multirow{2}{*}{Model} 
        & \multicolumn{3}{c}{ROUGE F1} 
        & \multicolumn{3}{c}{ROUGE Recall} 
        & \multirow{2}{*}{BERTScore-F1} 
        & \multirow{2}{*}{Training Hours} \\
        \cmidrule(lr){2-4} \cmidrule(lr){5-7}
        & R-1 & R-2 & R-L 
        & R-1 & R-2 & R-L 
        & & \\
        \midrule
        M-7B & 22.55 & 8.89 & 11.28 & 13.48 & 5.29 & 16.73 & 83.79 & -- \\
        SEA-E & 34.29 & 11.71 & 14.98 & 23.28& 7.90 & 10.13& 84.04 & -- \\
        SEA-EA & 36.97 & 13.02 & 15.88 & 25.62 & 8.97 & 10.97 & 84.15 & -- \\
        \midrule
        Mistral \((5, 3)\) & 53.44 & 15.47 & 22.39 & 53.18 & 15.42 & 22.31 & 86.44 & 4.34 \\
        Llama \((5, 3)\)& \underline{54.24} & \underline{15.91} & 22.59 & \textbf{53.75} & 15.77 & 22.41 & \underline{86.59} & 3.60 \\
        Mistral \((5, 5)\) & 53.65 & 15.80 & \underline{22.74} & \underline{53.60} & \underline{15.79} & \textbf{22.74} & 86.55 & 6.44 \\
        Llama \((5, 5)\) & \textbf{54.47} & \textbf{16.26} & \textbf{22.87} & 53.49 & \textbf{15.98} & \underline{22.48} & \textbf{86.68} & 5.30 \\
        \bottomrule
    \end{tabular}
\caption{Performance on the ICLR-24 test set comparing our models with baselines (M-7B, SEA-E, SEA-EA) from \citet{yu-etal-2024-automated}. \textbf{Bold} and \underline{underlined} values indicate best and second-best scores, respectively. Please note that the GPU configurations differ between Llama and Mistral fine-tuning (details in Appendix~\ref{sec:appendix-ft}).}
    \label{results:ICLR}
\end{table*}
This section presents a comprehensive evaluation of AutoRev across automatic, human, and robustness-focused analyses. We first benchmark feedback quality on the ICLR 2024 test set using ROUGE-1/2/L (Recall and F1) \cite{lin-2004-rouge} and BERTScore-F1 \cite{Zhang*2020BERTScore:} against the sections in the reference feedback, assessing both lexical overlap and semantic similarity. To evaluate feedback quality beyond surface similarity, we additionally report Aspect Coverage (ACov) \cite{10.1613/jair.1.12862}, which measures the breadth of scholarly aspects addressed in the generated feedback. Given the limitations of automated metrics, we conduct human evaluations assessing qualitative dimensions such as feedback depth, constructiveness, thoroughness, and helpfulness, alongside a distinct assessment of factual validity of the generated feedback. We further examine cross-venue generalization on NeurIPS and COLM 2025 papers and assess practical utility via an impact-oriented user survey. Finally, we employ a scalable LLM-as-a-Judge framework and perform ablation studies to analyze the effects of passage-selection parameters.



\subsection{Performance Comparison with Baselines}
\label{comparison_with_baseline}
We compare AutoRev against SEA-based baselines from \cite{yu-etal-2024-automated}, including M-7B, SEA-E, and SEA-EA. All baselines operate on the full paper text during both training and inference, without any learned passage selection. M-7B performs direct inference using an instruction-tuned Mistral model, SEA-E leverages standardized fine-tuning to improve review quality, and SEA-EA further incorporates a mismatch score to assess consistency between paper content and generated review.

In contrast, AutoRev employs learned passage selection using a Graph Neural Network over a document graph, providing only salient passages to the LLM. As shown in Table~\ref{results:ICLR}
, AutoRev consistently outperforms all baselines across ROUGE (Recall and F1) and BERTScore metrics. Our best-performing configuration, Llama \((5, 5)\) , achieves an average improvement of \textbf{58.72\%} over the state-of-the-art baselines.

To isolate the effects of model capacity from retrieval quality, we also evaluate AutoRev using Mistral. The substantial gains observed with Mistral confirm that the performance improvements stem primarily from GNN-based passage selection rather than the choice of LLM. The graph-based retrieval enables the model to capture both hierarchical and sequential dependencies across the document, providing the focused context necessary for generating comprehensive feedback.

Finally, Llama \((5, 3)\) offers a strong efficiency-performance trade-off, reducing training time by 32.07\% with only a 0.71\% decrease in performance relative to Llama \((5, 5)\) , while still outperforming all baselines.

\subsection{Aspect Coverage Analysis}
\label{aspect-coverage-analysis}
We evaluate the breadth of scholarly criteria addressed in the generated structured feedback using Aspect Coverage (ACov), following the aspect tagging framework of \citet{10.1613/jair.1.12862}. Unlike surface-level similarity metrics, ACov measures whether feedback engages with the diverse dimensions typically considered in peer review.

Table \ref{aspect-coverage-table} reports aspect coverage for our two best-performing configurations, Llama \((5, 3)\)  and Llama \((5, 5)\). Both variants demonstrate consistently strong coverage across core feedback dimensions, except for Motivation/Impact and Replicability, indicating that AutoRev generates feedback that goes beyond summarization and engages with substantive and technical aspects of the paper.

Overall, these results suggest that AutoRev provides broad and balanced coverage of evaluation criteria, reinforcing its suitability as a comprehensive pre-submission feedback system.


\begin{table*}[htbp]
  \centering
  \begin{tabular}{lcccccccc}
    \toprule
    \textbf{Model} & \textbf{SUM} & \textbf{CLA} & \textbf{SOU} & \textbf{SUB} & \textbf{ORI} & \textbf{CMP} & \textbf{REP} & \textbf{MOT} \\
    \midrule
    Llama \((5, 3)\) & 100.0 & 99.6 & 98.9 & 95.2 & 92.6 & 79.0 & 72.0 & 66.8 \\
    Llama \((5, 5)\) & 100.0 & 99.3 & 99.3 & 95.9 & 90.0 & 84.5 & 71.6 & 66.8 \\
    \bottomrule
  \end{tabular}
\caption{
\textbf{Aspect Coverage (\%)} of model-generated structured feedback evaluated using the Aspect Tagger \citep{10.1613/jair.1.12862} across aspects:
\textbf{Summary (SUM)}, \textbf{Clarity (CLA)}, \textbf{Soundness/Correctness (SOU)}, 
\textbf{Substance (SUB)}, \textbf{Originality (ORI)}, 
\textbf{Meaningful Comparison (CMP)}, \textbf{Replicability (REP)}, and \textbf{Motivation/Impact (MOT)}.
}
\label{aspect-coverage-table}
\end{table*}


\subsection{Human Evaluation}
\label{human-evaluation}
Understanding the limitations of automatic evaluation metrics, as highlighted by \citet{hanna-bojar-2021-fine} and \citet{schluter-2017-limits}, we conduct human evaluations to assess the acceptability, factual grounding, and practical utility of the generated feedback. Our evaluation is structured around two complementary dimensions: (i) qualitative assessment of feedback quality and (ii) factual validity of individual claims. We perform these evaluations on both the ICLR 2024 test set and a small out-of-domain dataset drawn from NeurIPS and COLM 2025. Due to the cost of expert annotation, human evaluations are conducted on representative subsets of the data. We report results for Llama \((5, 3)\) and Llama \((5, 5)\), our best-performing configurations.

\subsubsection{Qualitative Assessment}
\label{sec:human-qualitative}
A total of 30 papers are randomly selected from the ICLR 2024 test set (ICT\_30). Each paper is associated with three feedback instances: one reference feedback corresponding to the ICT\_30 paper and one generated by each of Llama \((5, 3)\) and Llama \((5, 5)\), yielding 90 feedback samples. A total of 27 annotators participate in the study, with each feedback annotated by three independent annotators.
Feedback quality is evaluated across four criteria:
\begin{enumerate}[noitemsep]
  \item \textit{Feedback Depth (FD)}: Evaluates the technical understanding and meaningful engagement with the paper's content.
  \item \textit{Feedback Thoroughness (FT)}: Measures the coverage of key components like methodology, results, and reproducibility.
  \item \textit{Feedback Constructiveness (FC)}: Assesses the utility of suggestions for improvement.
  \item \textit{Feedback Helpfulness (FH)}: Evaluates how effectively important issues are identified and explained to the authors.
\end{enumerate}
Annotators assign scores on a 1-5 scale following the guidelines detailed in Appendix \ref{app:llm-as-a-judge} and snapshots of the annotation interface are shown in Appendix \ref{app:human_eval_ui}. Scores are averaged across annotators for each feedback and then aggregated across all samples for each model.

As shown in Table \ref{tab:quality_assessment}, AutoRev-generated feedback closely approaches ICT\_30 feedback in constructiveness and helpfulness, which are particularly critical for pre-submission support. While ICT\_30 feedback remains superior in depth and thoroughness, the results show that AutoRev does not merely summarize paper content, but meaningfully engages with methodological and clarity issues, reinforcing its suitability as a structured feedback tool for authors.


\begin{table}[htbp]
  \centering
  \begin{tabular}{lcccc}
    \toprule
    \textbf{Feedback} & \textbf{FD} & \textbf{FT} & \textbf{FC} & \textbf{FH} \\
    \midrule
    ICT\_30 & 3.54 & 3.49 & 3.26 & 3.41 \\
    \midrule
    Llama \((5, 3)\) & 3.37 & 3.30 & 3.27 & 3.34 \\
    Llama \((5, 5)\) & 3.27 & 3.24 & 3.19 & 3.30 \\
    \bottomrule
  \end{tabular}
  \caption{Human Qualitative Assessment scores (scale of 1--5) of generated feedback using Llama models against reference feedback corresponding to the ICT\_30 paper across four criteria: Feedback Depth (FD), Feedback Thoroughness (FT), Feedback Constructiveness (FC), and Feedback Helpfulness (FH).}
  \label{tab:quality_assessment}
\end{table}

\subsubsection{Factual Validity Assessment}
\label{sec:factual}
To evaluate factual grounding, we randomly select 10 papers from the ICLR 2024 test set (ICT\_10). Each paper is associated with three feedback instances: one reference feedback corresponding to the ICT\_10 paper and outputs from Llama \((5, 3)\) and Llama \((5, 5)\). These feedbacks are annotated by 10 annotators, with each paper evaluated by two annotators.

Annotators assess each statement within a feedback based on its factual consistency with the source paper, using a four-point scale: Completely Agree, Mostly Agree, Mostly Disagree, and Completely Disagree. The evaluation guidelines are provided in Appendix \ref{app:eval_guidelines_human}, and snapshots of the annotation interface are shown in Appendix \ref{app:human_eval_ui}.

The percentage agreement scores summarized in Tables \ref{tab:indiv_rating_percent} and \ref{tab:pos_rating} indicate that AutoRev maintains strong factual grounding in the source document. While ICT\_10 feedback achieves the highest agreement, AutoRev-generated feedback receives a clear majority of positive ratings, with complete disagreements (potential hallucinations) remaining below 15\%. These findings suggest that graph-based passage selection effectively anchors feedback generation in relevant evidence, mitigating hallucinated content.



\subsubsection{Cross-Venue Generalization Assessment}
\label{sec:out-of-domain}

To further examine robustness beyond the training distribution, we conduct an out-of-domain human evaluation on a curated set of 10 papers, comprising five papers each from COLM 2025 and NeurIPS 2025 (CNT\_10). Each paper is evaluated using feedback generated by Llama \((5, 3)\) and Llama \((5, 5)\). Due to the absence of publicly available consolidated reference feedback for these venues and the high cost of expert annotation, reference feedback is not included.

A total of 10 annotators participate in the study, with each paper evaluated by two annotators following the same guidelines described in Section~\ref{sec:factual}. The results, reported in Table~\ref{tab:indiv_rating_percent} and \ref{tab:pos_rating}, show comparable factual agreement across venues. This suggests that AutoRev generalizes beyond the ICLR 2024 training distribution, capturing structural and semantic regularities common to scientific writing rather than overfitting to venue-specific conventions.


\begin{table}[htbp]
  \centering
  \setlength{\tabcolsep}{2pt} 
  \begin{tabular}{llcccc}
    \toprule
    \textbf{Dataset} & \textbf{Feedback} & \textbf{CA} & \textbf{MA} & \textbf{MD} & \textbf{CD} \\
    \midrule
    \multirow{3}{*}{\shortstack[l]{ICT\_10}} 
      & Reference & 39.12 & 37.85 & 16.40 & 6.62 \\
      & Llama \((5, 3)\) & 32.64 & 31.94 & 22.92 & 12.50 \\
      & Llama \((5, 5)\) & 32.27 & 32.98 & 23.05 & 11.70 \\
    \midrule
    \multirow{2}{*}{CNT\_10} 
      & Llama \((5, 3)\) & 22.00 & 39.86 & 24.57 & 13.57 \\
      & Llama \((5, 5)\) & 24.06 & 40.25 & 22.19 & 13.50 \\
    \bottomrule
  \end{tabular}
  \caption{Factual validity assessment (\% agreement) on ICLR 2024 (ICT\_10) and cross-venue CNT\_10 datasets.
CA/MA indicate grounded statements, while CD reflects clear factual errors (hallucinations).}
  \label{tab:indiv_rating_percent}
\end{table}

\subsubsection{Impact Assessment}
\label{sec:impact}
To evaluate the impact of our system, we conduct a survey among the annotators involved in the tasks discussed in Sections~\ref{sec:factual} and \ref{sec:out-of-domain}. Participants are asked: \textit{``As an author, how likely are you to use AutoRev to get feedback for your own manuscript before submitting it to a conference?''} On a 5-point scale, the system received an average rating of \textbf{4.2}. This indicates that authors are strongly inclined to use our model for pre-submission critique, demonstrating the system's practical utility.
\subsection{LLM-as-a-Judge Evaluation}
\label{llm-as-judge}
As mentioned by \citet{10.5555/3666122.3668142}, the LLM-as-a-Judge method offers an explainable and affordable way to approximate human preferences, which are otherwise costly to gather. 
We employ a large language model, Gemini-2.0-Flash\footnote{\url{https://deepmind.google/technologies/gemini/}} as an automated judge to evaluate generated feedback. Since LLM-as-a-Judge evaluation is scalable and much cheaper to undertake, we evaluate all our models against the reference feedback, over the entire ICT dataset. The evaluation criteria and prompting guidelines are adapted from those used in our human qualitative assessment (Section~\ref{sec:human-qualitative}; Appendix~\ref{app:llm-as-a-judge}). The results of this experiment are summarized in Table~\ref{llmjudge}

A key finding is the strong correlation between LLM-as-a-Judge scores and human annotations (Table~\ref{tab:quality_assessment}). This alignment validates the automated judge as a reliable proxy for assessing feedback quality in this domain.
\begin{table}[htbp]
    \centering
    \begin{tabular}{lcccc}
        \toprule
        \textbf{Feedback} 
         & \textbf{FD} & \textbf{FT} & \textbf{FC} & \textbf{FH} \\
        \midrule
        ICT & 3.98 & 3.87 & 3.79 &3.79  \\
        \midrule
        Mistral \((5, 3)\) & 3.50 & 3.35 & 3.44 & 3.32  \\
        Llama \((5, 3)\) & \underline{3.61} & \textbf{3.45} & \textbf{3.47} & \underline{3.39}  \\
        Mistral \((5, 5)\) & 3.48 & 3.36 & 3.43 & 3.34 \\
        Llama \((5, 5)\) & \textbf{3.63} & \underline{3.44} & \textbf{3.47} & \textbf{3.41}  \\
        \bottomrule
    \end{tabular}
\caption{LLM-as-a-Judge (Gemini-2.0-Flash) scores (scale of 1--5) comparing model-generated feedback against ICT across Feedback Depth (FD), Feedback Thoroughness (FT), Feedback Constructiveness (FC), and Feedback Helpfulness (FH). \textbf{Bold} and \underline{underlined} values indicate the best and second-best model scores (excluding reference feedback), respectively.}
\label{llmjudge}
\end{table}
\subsection{Ablation Studies} 
\label{ablation-studies-paper} 
We perform ablation studies by systematically varying the values of \(k \in \{1, 3, 5\}\) and \(m \in \{3, 5\}\) during LLM fine-tuning. The detailed results are reported in Appendix~\ref{sec:abl-studies}. The results indicate that increasing \(k\) and \(m\) provides richer context to both Mistral and Llama to generate feedback, leading to consistent improvements across all evaluation metrics. Though, this comes at the downside of higher training times as \(m\) increases for a fixed \(k\).

\section{Conclusion}
In this work, we present AutoRev, an author-centric system for generating structured pre-submission feedback using graph-based passage retrieval. By modeling papers as hierarchical document graphs and leveraging a GNN to identify salient content, AutoRev substantially reduces LLM input length, overcoming the limitations of full-paper fine-tuning, stemming from long-context degradation \cite{hsieh2024ruler}, while operating without human-written grounding. Experiments on ICLR 2024 show that AutoRev demonstrates strong empirical performance through automatic metrics, with human evaluations confirming constructive, well-grounded feedback and cross-venue generalization. Notably, Llama $(5, 5)$ excels in generating comprehensive feedback for rigorous refinement of manuscripts, whereas Llama $(5, 3)$ offers a favorable efficiency-performance trade-off for large-scale deployment.




\section*{Limitations}
While our framework provides a robust and high-performing solution for feedback generation, we acknowledge that no automatic system can fully replicate human judgment in complex tasks such as peer reviewing, as noted by \citet{ye2024yetrevealingrisksutilizing}. Our goal is to assist authors in refining their papers and we aim for our system to support authors only in the pre-submission feedback phase. Currently, our system is primarily focused on ICLR 2024 papers due to the limited availability of public datasets containing consolidated reviews. However, the consistent positive performance observed in out-of-domain human evaluations (Section~\ref{sec:out-of-domain}) suggests that our model generalizes well beyond the ICLR 2024 setting.

Another limitation is our reliance on human-written reviews consolidated using GPT-4. While consolidated reviews provide a clearer and more consistent supervision signal than multi-review datasets, which are often noisy and contradictory and can hinder model learning, this approach introduces its own challenges. In particular, relying on an LLM for consolidation may induce biases. Future work can explore hybrid strategies that reduce or eliminate reliance on LLMs for consolidating multiple reviews into a single supervision signal.


\section*{Ethical Considerations}
The deployment of AutoRev raises ethical concerns regarding bias, misuse, and academic integrity. Critically, while the system is designed strictly as an author-centric aid, there is a risk of over-reliance or misuse in decision-making, which could undermine the nuances of human reviewing. Furthermore, by training on historical peer reviews and GPT-4 consolidated data, the model may inherit and amplify existing biases present in those texts, such as leniency toward well-known authors or under-representation of certain research areas, potentially skewing evaluations. Additionally, the high fidelity of the generated text raises concerns about academic misconduct, where AI critiques could be misrepresented as original human work. To address these issues, we advocate for transparency and emphasize that AutoRev is intended solely as a supplementary tool to support, not replace, the scientific review process.

\bibliography{custom}

@String{Academic = "Academic Press" }

@String{Chelsea = "Chelsea" }

@String{Springer = "Springer-Verlag" }

@inproceedings{zhao-etal-2024-hierarchical,
    title = "Hierarchical Attention Graph for Scientific Document Summarization in Global and Local Level",
    author = "Zhao, Chenlong  and
      Zhou, Xiwen  and
      Xie, Xiaopeng  and
      Zhang, Yong",
    editor = "Duh, Kevin  and
      Gomez, Helena  and
      Bethard, Steven",
    booktitle = "Findings of the Association for Computational Linguistics: NAACL 2024",
    month = jun,
    year = "2024",
    address = "Mexico City, Mexico",
    publisher = "Association for Computational Linguistics",
    url = "https://aclanthology.org/2024.findings-naacl.45/",
    doi = "10.18653/v1/2024.findings-naacl.45",
    pages = "714--726"
}

@inproceedings{bird-loper-2004-nltk,
    title = "{NLTK}: The Natural Language Toolkit",
    author = "Bird, Steven  and
      Loper, Edward",
    booktitle = "Proceedings of the {ACL} Interactive Poster and Demonstration Sessions",
    month = jul,
    year = "2004",
    address = "Barcelona, Spain",
    publisher = "Association for Computational Linguistics",
    url = "https://aclanthology.org/P04-3031/",
    pages = "214--217"
}

@inproceedings{
veličković2018graph,
title={Graph Attention Networks},
author={Petar Veličković and Guillem Cucurull and Arantxa Casanova and Adriana Romero and Pietro Liò and Yoshua Bengio},
booktitle={International Conference on Learning Representations},
year={2018},
url={https://openreview.net/forum?id=rJXMpikCZ},
}

@InProceedings{10.1007/978-3-031-70344-7_7,
author="Faizullah, Abdur Rahman Bin Mohammed
and Urlana, Ashok
and Mishra, Rahul",
editor="Bifet, Albert
and Davis, Jesse
and Krilavi{\v{c}}ius, Tomas
and Kull, Meelis
and Ntoutsi, Eirini
and {\v{Z}}liobait{\.{e}}, Indr{\.{e}}",
title="LimGen: Probing the LLMs for Generating Suggestive Limitations of Research Papers",
booktitle="Machine Learning and Knowledge Discovery in Databases. Research Track",
year="2024",
publisher="Springer Nature Switzerland",
address="Cham",
pages="106--124",
isbn="978-3-031-70344-7"
}

@article{10.1613/jair.1.12862,
author = {Yuan, Weizhe and Liu, Pengfei and Neubig, Graham},
title = {Can We Automate Scientific Reviewing?},
year = {2022},
issue_date = {Dec 2022},
publisher = {AI Access Foundation},
address = {El Segundo, CA, USA},
volume = {75},
issn = {1076-9757},
url = {https://doi.org/10.1613/jair.1.12862},
doi = {10.1613/jair.1.12862},
journal = {J. Artif. Int. Res.},
month = dec,
numpages = {42}
}

@inproceedings{yu-etal-2024-automated,
    title = "Automated Peer Reviewing in Paper {SEA}: Standardization, Evaluation, and Analysis",
    author = "Yu, Jianxiang  and
      Ding, Zichen  and
      Tan, Jiaqi  and
      Luo, Kangyang  and
      Weng, Zhenmin  and
      Gong, Chenghua  and
      Zeng, Long  and
      Cui, RenJing  and
      Han, Chengcheng  and
      Sun, Qiushi  and
      Wu, Zhiyong  and
      Lan, Yunshi  and
      Li, Xiang",
    editor = "Al-Onaizan, Yaser  and
      Bansal, Mohit  and
      Chen, Yun-Nung",
    booktitle = "Findings of the Association for Computational Linguistics: EMNLP 2024",
    month = nov,
    year = "2024",
    address = "Miami, Florida, USA",
    publisher = "Association for Computational Linguistics",
    url = "https://aclanthology.org/2024.findings-emnlp.595/",
    doi = "10.18653/v1/2024.findings-emnlp.595",
    pages = "10164--10184"
}

@inproceedings{idahl-ahmadi-2025-openreviewer,
    title = "{O}pen{R}eviewer: A Specialized Large Language Model for Generating Critical Scientific Paper Reviews",
    author = "Idahl, Maximilian  and
      Ahmadi, Zahra",
    editor = "Dziri, Nouha  and
      Ren, Sean (Xiang)  and
      Diao, Shizhe",
    booktitle = "Proceedings of the 2025 Conference of the Nations of the Americas Chapter of the Association for Computational Linguistics: Human Language Technologies (System Demonstrations)",
    month = apr,
    year = "2025",
    address = "Albuquerque, New Mexico",
    publisher = "Association for Computational Linguistics",
    url = "https://aclanthology.org/2025.naacl-demo.44/",
    doi = "10.18653/v1/2025.naacl-demo.44",
    pages = "550--562",
    ISBN = "979-8-89176-191-9"
}

@misc{darcy2024margmultiagentreviewgeneration,
      title={MARG: Multi-Agent Review Generation for Scientific Papers}, 
      author={Mike D'Arcy and Tom Hope and Larry Birnbaum and Doug Downey},
      year={2024},
      eprint={2401.04259},
      archivePrefix={arXiv},
      primaryClass={cs.CL},
      url={https://arxiv.org/abs/2401.04259}, 
}

@misc{zhu2025graphbasedapproachesfunctionalitiesretrievalaugmented,
      title={Graph-based Approaches and Functionalities in Retrieval-Augmented Generation: A Comprehensive Survey}, 
      author={Zulun Zhu and Tiancheng Huang and Kai Wang and Junda Ye and Xinghe Chen and Siqiang Luo},
      year={2025},
      eprint={2504.10499},
      archivePrefix={arXiv},
      primaryClass={cs.IR},
      url={https://arxiv.org/abs/2504.10499}, 
}

@inproceedings{10.5555/3495724.3496517,
author = {Lewis, Patrick and Perez, Ethan and Piktus, Aleksandra and Petroni, Fabio and Karpukhin, Vladimir and Goyal, Naman and K\"{u}ttler, Heinrich and Lewis, Mike and Yih, Wen-tau and Rockt\"{a}schel, Tim and Riedel, Sebastian and Kiela, Douwe},
title = {Retrieval-augmented generation for knowledge-intensive NLP tasks},
year = {2020},
isbn = {9781713829546},
publisher = {Curran Associates Inc.},
address = {Red Hook, NY, USA},
booktitle = {Proceedings of the 34th International Conference on Neural Information Processing Systems},
articleno = {793},
numpages = {16},
location = {Vancouver, BC, Canada},
series = {NIPS '20}
}

@inproceedings{reimers-gurevych-2019-sentence,
    title = "Sentence-{BERT}: Sentence Embeddings using {S}iamese {BERT}-Networks",
    author = "Reimers, Nils  and
      Gurevych, Iryna",
    editor = "Inui, Kentaro  and
      Jiang, Jing  and
      Ng, Vincent  and
      Wan, Xiaojun",
    booktitle = "Proceedings of the 2019 Conference on Empirical Methods in Natural Language Processing and the 9th International Joint Conference on Natural Language Processing (EMNLP-IJCNLP)",
    month = nov,
    year = "2019",
    address = "Hong Kong, China",
    publisher = "Association for Computational Linguistics",
    url = "https://aclanthology.org/D19-1410/",
    doi = "10.18653/v1/D19-1410",
    pages = "3982--3992"
}

@inproceedings{chitale-etal-2025-discograms,
    title = "{D}isco{G}ra{MS}: Enhancing Movie Screen-Play Summarization using Movie Character-Aware Discourse Graph",
    author = "Chitale, Maitreya Prafulla  and
      Bindal, Uday  and
      Rajkumar, Rajakrishnan P  and
      Mishra, Rahul",
    editor = "Chiruzzo, Luis  and
      Ritter, Alan  and
      Wang, Lu",
    booktitle = "Proceedings of the 2025 Conference of the Nations of the Americas Chapter of the Association for Computational Linguistics: Human Language Technologies (Volume 2: Short Papers)",
    month = apr,
    year = "2025",
    address = "Albuquerque, New Mexico",
    publisher = "Association for Computational Linguistics",
    url = "https://aclanthology.org/2025.naacl-short.80/",
    pages = "954--965",
    ISBN = "979-8-89176-190-2"
}

@inproceedings{schluter-2017-limits,
    title = "The limits of automatic summarisation according to {ROUGE}",
    author = "Schluter, Natalie",
    editor = "Lapata, Mirella  and
      Blunsom, Phil  and
      Koller, Alexander",
    booktitle = "Proceedings of the 15th Conference of the {E}uropean Chapter of the Association for Computational Linguistics: Volume 2, Short Papers",
    month = apr,
    year = "2017",
    address = "Valencia, Spain",
    publisher = "Association for Computational Linguistics",
    url = "https://aclanthology.org/E17-2007/",
    pages = "41--45"
}

@inproceedings{hanna-bojar-2021-fine,
    title = "A Fine-Grained Analysis of {BERTS}core",
    author = "Hanna, Michael  and
      Bojar, Ond{\v{r}}ej",
    editor = "Barrault, Loic  and
      Bojar, Ondrej  and
      Bougares, Fethi  and
      Chatterjee, Rajen  and
      Costa-jussa, Marta R.  and
      Federmann, Christian  and
      Fishel, Mark  and
      Fraser, Alexander  and
      Freitag, Markus  and
      Graham, Yvette  and
      Grundkiewicz, Roman  and
      Guzman, Paco  and
      Haddow, Barry  and
      Huck, Matthias  and
      Yepes, Antonio Jimeno  and
      Koehn, Philipp  and
      Kocmi, Tom  and
      Martins, Andre  and
      Morishita, Makoto  and
      Monz, Christof",
    booktitle = "Proceedings of the Sixth Conference on Machine Translation",
    month = nov,
    year = "2021",
    address = "Online",
    publisher = "Association for Computational Linguistics",
    url = "https://aclanthology.org/2021.wmt-1.59/",
    pages = "507--517"
}

@inproceedings{gao-etal-2025-mixed,
    title = "A Mixed-Language Multi-Document News Summarization Dataset and a Graphs-Based Extract-Generate Model",
    author = "Gao, Shengxiang  and
      Nan, Fang  and
      Zhang, Yongbing  and
      Huang, Yuxin  and
      Tan, Kaiwen  and
      Yu, Zhengtao",
    editor = "Chiruzzo, Luis  and
      Ritter, Alan  and
      Wang, Lu",
    booktitle = "Proceedings of the 2025 Conference of the Nations of the Americas Chapter of the Association for Computational Linguistics: Human Language Technologies (Volume 1: Long Papers)",
    month = apr,
    year = "2025",
    address = "Albuquerque, New Mexico",
    publisher = "Association for Computational Linguistics",
    url = "https://aclanthology.org/2025.naacl-long.468/",
    pages = "9255--9265",
    ISBN = "979-8-89176-189-6"
}

@inproceedings{agarwal-etal-2025-hybrid,
    title = "Hybrid Graphs for Table-and-Text based Question Answering using {LLM}s",
    author = "Agarwal, Ankush  and
      Devaguptapu, Chaitanya  and
      S, Ganesh",
    editor = "Chiruzzo, Luis  and
      Ritter, Alan  and
      Wang, Lu",
    booktitle = "Proceedings of the 2025 Conference of the Nations of the Americas Chapter of the Association for Computational Linguistics: Human Language Technologies (Volume 1: Long Papers)",
    month = apr,
    year = "2025",
    address = "Albuquerque, New Mexico",
    publisher = "Association for Computational Linguistics",
    url = "https://aclanthology.org/2025.naacl-long.39/",
    pages = "858--875",
    ISBN = "979-8-89176-189-6"
}

@inproceedings{aitymbetov-zorbas-2025-autonomous,
    title = "Autonomous Machine Learning-Based Peer Reviewer Selection System",
    author = "Aitymbetov, Nurmukhammed  and
      Zorbas, Dimitrios",
    editor = "Rambow, Owen  and
      Wanner, Leo  and
      Apidianaki, Marianna  and
      Al-Khalifa, Hend  and
      Eugenio, Barbara Di  and
      Schockaert, Steven  and
      Mather, Brodie  and
      Dras, Mark",
    booktitle = "Proceedings of the 31st International Conference on Computational Linguistics: System Demonstrations",
    month = jan,
    year = "2025",
    address = "Abu Dhabi, UAE",
    publisher = "Association for Computational Linguistics",
    url = "https://aclanthology.org/2025.coling-demos.20/",
    pages = "199--207"
}

@misc{ye2024yetrevealingrisksutilizing,
      title={Are We There Yet? Revealing the Risks of Utilizing Large Language Models in Scholarly Peer Review}, 
      author={Rui Ye and Xianghe Pang and Jingyi Chai and Jiaao Chen and Zhenfei Yin and Zhen Xiang and Xiaowen Dong and Jing Shao and Siheng Chen},
      year={2024},
      eprint={2412.01708},
      archivePrefix={arXiv},
      primaryClass={cs.CL},
      url={https://arxiv.org/abs/2412.01708}, 
}

@inproceedings{karpukhin-etal-2020-dense,
    title = "Dense Passage Retrieval for Open-Domain Question Answering",
    author = "Karpukhin, Vladimir  and
      Oguz, Barlas  and
      Min, Sewon  and
      Lewis, Patrick  and
      Wu, Ledell  and
      Edunov, Sergey  and
      Chen, Danqi  and
      Yih, Wen-tau",
    editor = "Webber, Bonnie  and
      Cohn, Trevor  and
      He, Yulan  and
      Liu, Yang",
    booktitle = "Proceedings of the 2020 Conference on Empirical Methods in Natural Language Processing (EMNLP)",
    month = nov,
    year = "2020",
    address = "Online",
    publisher = "Association for Computational Linguistics",
    url = "https://aclanthology.org/2020.emnlp-main.550/",
    doi = "10.18653/v1/2020.emnlp-main.550",
    pages = "6769--6781"
}

@inproceedings{liebling-etal-2025-towards,
    title = "Towards {AI}-assisted Academic Writing",
    author = "Liebling, Daniel J.  and
      Kane, Malcolm  and
      Grunde-McLaughlin, Madeleine  and
      Lang, Ian  and
      Venugopalan, Subhashini  and
      Brenner, Michael",
    editor = "Jansen, Peter  and
      Dalvi Mishra, Bhavana  and
      Trivedi, Harsh  and
      Prasad Majumder, Bodhisattwa  and
      Hope, Tom  and
      Khot, Tushar  and
      Downey, Doug  and
      Horvitz, Eric",
    booktitle = "Proceedings of the 1st Workshop on AI and Scientific Discovery: Directions and Opportunities",
    month = may,
    year = "2025",
    address = "Albuquerque, New Mexico, USA",
    publisher = "Association for Computational Linguistics",
    url = "https://aclanthology.org/2025.aisd-main.4/",
    doi = "10.18653/v1/2025.aisd-main.4",
    pages = "31--45",
    ISBN = "979-8-89176-224-4"
}

@inproceedings{hu-wan-2014-automatic,
    title = "Automatic Generation of Related Work Sections in Scientific Papers: An Optimization Approach",
    author = "Hu, Yue  and
      Wan, Xiaojun",
    editor = "Moschitti, Alessandro  and
      Pang, Bo  and
      Daelemans, Walter",
    booktitle = "Proceedings of the 2014 Conference on Empirical Methods in Natural Language Processing ({EMNLP})",
    month = oct,
    year = "2014",
    address = "Doha, Qatar",
    publisher = "Association for Computational Linguistics",
    url = "https://aclanthology.org/D14-1170/",
    doi = "10.3115/v1/D14-1170",
    pages = "1624--1633"
}

@inproceedings{gu-hahnloser-2023-scilit,
    title = "{S}ci{L}it: A Platform for Joint Scientific Literature Discovery, Summarization and Citation Generation",
    author = "Gu, Nianlong  and
      Hahnloser, Richard H.R.",
    editor = "Bollegala, Danushka  and
      Huang, Ruihong  and
      Ritter, Alan",
    booktitle = "Proceedings of the 61st Annual Meeting of the Association for Computational Linguistics (Volume 3: System Demonstrations)",
    month = jul,
    year = "2023",
    address = "Toronto, Canada",
    publisher = "Association for Computational Linguistics",
    url = "https://aclanthology.org/2023.acl-demo.22/",
    doi = "10.18653/v1/2023.acl-demo.22",
    pages = "235--246"
}

@misc{openai2024gpt4technicalreport,
      title={GPT-4 Technical Report}, 
      author={OpenAI and Josh Achiam and Steven Adler and Sandhini Agarwal and Lama Ahmad and Ilge Akkaya and Florencia Leoni Aleman and Diogo Almeida and Janko Altenschmidt and Sam Altman and Shyamal Anadkat and Red Avila and Igor Babuschkin and Suchir Balaji and Valerie Balcom and Paul Baltescu and Haiming Bao and Mohammad Bavarian and Jeff Belgum and Irwan Bello and Jake Berdine and Gabriel Bernadett-Shapiro and Christopher Berner and Lenny Bogdonoff and Oleg Boiko and Madelaine Boyd and Anna-Luisa Brakman and Greg Brockman and Tim Brooks and Miles Brundage and Kevin Button and Trevor Cai and Rosie Campbell and Andrew Cann and Brittany Carey and Chelsea Carlson and Rory Carmichael and Brooke Chan and Che Chang and Fotis Chantzis and Derek Chen and Sully Chen and Ruby Chen and Jason Chen and Mark Chen and Ben Chess and Chester Cho and Casey Chu and Hyung Won Chung and Dave Cummings and Jeremiah Currier and Yunxing Dai and Cory Decareaux and Thomas Degry and Noah Deutsch and Damien Deville and Arka Dhar and David Dohan and Steve Dowling and Sheila Dunning and Adrien Ecoffet and Atty Eleti and Tyna Eloundou and David Farhi and Liam Fedus and Niko Felix and Simón Posada Fishman and Juston Forte and Isabella Fulford and Leo Gao and Elie Georges and Christian Gibson and Vik Goel and Tarun Gogineni and Gabriel Goh and Rapha Gontijo-Lopes and Jonathan Gordon and Morgan Grafstein and Scott Gray and Ryan Greene and Joshua Gross and Shixiang Shane Gu and Yufei Guo and Chris Hallacy and Jesse Han and Jeff Harris and Yuchen He and Mike Heaton and Johannes Heidecke and Chris Hesse and Alan Hickey and Wade Hickey and Peter Hoeschele and Brandon Houghton and Kenny Hsu and Shengli Hu and Xin Hu and Joost Huizinga and Shantanu Jain and Shawn Jain and Joanne Jang and Angela Jiang and Roger Jiang and Haozhun Jin and Denny Jin and Shino Jomoto and Billie Jonn and Heewoo Jun and Tomer Kaftan and Łukasz Kaiser and Ali Kamali and Ingmar Kanitscheider and Nitish Shirish Keskar and Tabarak Khan and Logan Kilpatrick and Jong Wook Kim and Christina Kim and Yongjik Kim and Jan Hendrik Kirchner and Jamie Kiros and Matt Knight and Daniel Kokotajlo and Łukasz Kondraciuk and Andrew Kondrich and Aris Konstantinidis and Kyle Kosic and Gretchen Krueger and Vishal Kuo and Michael Lampe and Ikai Lan and Teddy Lee and Jan Leike and Jade Leung and Daniel Levy and Chak Ming Li and Rachel Lim and Molly Lin and Stephanie Lin and Mateusz Litwin and Theresa Lopez and Ryan Lowe and Patricia Lue and Anna Makanju and Kim Malfacini and Sam Manning and Todor Markov and Yaniv Markovski and Bianca Martin and Katie Mayer and Andrew Mayne and Bob McGrew and Scott Mayer McKinney and Christine McLeavey and Paul McMillan and Jake McNeil and David Medina and Aalok Mehta and Jacob Menick and Luke Metz and Andrey Mishchenko and Pamela Mishkin and Vinnie Monaco and Evan Morikawa and Daniel Mossing and Tong Mu and Mira Murati and Oleg Murk and David Mély and Ashvin Nair and Reiichiro Nakano and Rajeev Nayak and Arvind Neelakantan and Richard Ngo and Hyeonwoo Noh and Long Ouyang and Cullen O'Keefe and Jakub Pachocki and Alex Paino and Joe Palermo and Ashley Pantuliano and Giambattista Parascandolo and Joel Parish and Emy Parparita and Alex Passos and Mikhail Pavlov and Andrew Peng and Adam Perelman and Filipe de Avila Belbute Peres and Michael Petrov and Henrique Ponde de Oliveira Pinto and Michael and Pokorny and Michelle Pokrass and Vitchyr H. Pong and Tolly Powell and Alethea Power and Boris Power and Elizabeth Proehl and Raul Puri and Alec Radford and Jack Rae and Aditya Ramesh and Cameron Raymond and Francis Real and Kendra Rimbach and Carl Ross and Bob Rotsted and Henri Roussez and Nick Ryder and Mario Saltarelli and Ted Sanders and Shibani Santurkar and Girish Sastry and Heather Schmidt and David Schnurr and John Schulman and Daniel Selsam and Kyla Sheppard and Toki Sherbakov and Jessica Shieh and Sarah Shoker and Pranav Shyam and Szymon Sidor and Eric Sigler and Maddie Simens and Jordan Sitkin and Katarina Slama and Ian Sohl and Benjamin Sokolowsky and Yang Song and Natalie Staudacher and Felipe Petroski Such and Natalie Summers and Ilya Sutskever and Jie Tang and Nikolas Tezak and Madeleine B. Thompson and Phil Tillet and Amin Tootoonchian and Elizabeth Tseng and Preston Tuggle and Nick Turley and Jerry Tworek and Juan Felipe Cerón Uribe and Andrea Vallone and Arun Vijayvergiya and Chelsea Voss and Carroll Wainwright and Justin Jay Wang and Alvin Wang and Ben Wang and Jonathan Ward and Jason Wei and CJ Weinmann and Akila Welihinda and Peter Welinder and Jiayi Weng and Lilian Weng and Matt Wiethoff and Dave Willner and Clemens Winter and Samuel Wolrich and Hannah Wong and Lauren Workman and Sherwin Wu and Jeff Wu and Michael Wu and Kai Xiao and Tao Xu and Sarah Yoo and Kevin Yu and Qiming Yuan and Wojciech Zaremba and Rowan Zellers and Chong Zhang and Marvin Zhang and Shengjia Zhao and Tianhao Zheng and Juntang Zhuang and William Zhuk and Barret Zoph},
      year={2024},
      eprint={2303.08774},
      archivePrefix={arXiv},
      primaryClass={cs.CL},
      url={https://arxiv.org/abs/2303.08774}, 
}

@misc{jiang2023mistral7b,
      title={Mistral 7B}, 
      author={Albert Q. Jiang and Alexandre Sablayrolles and Arthur Mensch and Chris Bamford and Devendra Singh Chaplot and Diego de las Casas and Florian Bressand and Gianna Lengyel and Guillaume Lample and Lucile Saulnier and Lélio Renard Lavaud and Marie-Anne Lachaux and Pierre Stock and Teven Le Scao and Thibaut Lavril and Thomas Wang and Timothée Lacroix and William El Sayed},
      year={2023},
      eprint={2310.06825},
      archivePrefix={arXiv},
      primaryClass={cs.CL},
      url={https://arxiv.org/abs/2310.06825}, 
}

@article{10.1007/s00799-023-00359-0,
author = {Kumar, Asheesh and Ghosal, Tirthankar and Bhattacharjee, Saprativa and Ekbal, Asif},
title = {Towards automated meta-review generation via an NLP/ML pipeline in different stages of the scholarly peer review process},
year = {2023},
issue_date = {Sep 2024},
publisher = {Springer-Verlag},
address = {Berlin, Heidelberg},
volume = {25},
number = {3},
issn = {1432-5012},
url = {https://doi.org/10.1007/s00799-023-00359-0},
doi = {10.1007/s00799-023-00359-0},
journal = {Int. J. Digit. Libr.},
month = apr,
pages = {493–504},
numpages = {12}
}

@misc{grattafiori2024llama3herdmodels,
      title={The Llama 3 Herd of Models}, 
      author={Aaron Grattafiori and Abhimanyu Dubey and Abhinav Jauhri and Abhinav Pandey and Abhishek Kadian and Ahmad Al-Dahle and Aiesha Letman and Akhil Mathur and Alan Schelten and Alex Vaughan and Amy Yang and Angela Fan and Anirudh Goyal and Anthony Hartshorn and Aobo Yang and Archi Mitra and Archie Sravankumar and Artem Korenev and Arthur Hinsvark and Arun Rao and Aston Zhang and Aurelien Rodriguez and Austen Gregerson and Ava Spataru and Baptiste Roziere and Bethany Biron and Binh Tang and Bobbie Chern and Charlotte Caucheteux and Chaya Nayak and Chloe Bi and Chris Marra and Chris McConnell and Christian Keller and Christophe Touret and Chunyang Wu and Corinne Wong and Cristian Canton Ferrer and Cyrus Nikolaidis and Damien Allonsius and Daniel Song and Danielle Pintz and Danny Livshits and Danny Wyatt and David Esiobu and Dhruv Choudhary and Dhruv Mahajan and Diego Garcia-Olano and Diego Perino and Dieuwke Hupkes and Egor Lakomkin and Ehab AlBadawy and Elina Lobanova and Emily Dinan and Eric Michael Smith and Filip Radenovic and Francisco Guzmán and Frank Zhang and Gabriel Synnaeve and Gabrielle Lee and Georgia Lewis Anderson and Govind Thattai and Graeme Nail and Gregoire Mialon and Guan Pang and Guillem Cucurell and Hailey Nguyen and Hannah Korevaar and Hu Xu and Hugo Touvron and Iliyan Zarov and Imanol Arrieta Ibarra and Isabel Kloumann and Ishan Misra and Ivan Evtimov and Jack Zhang and Jade Copet and Jaewon Lee and Jan Geffert and Jana Vranes and Jason Park and Jay Mahadeokar and Jeet Shah and Jelmer van der Linde and Jennifer Billock and Jenny Hong and Jenya Lee and Jeremy Fu and Jianfeng Chi and Jianyu Huang and Jiawen Liu and Jie Wang and Jiecao Yu and Joanna Bitton and Joe Spisak and Jongsoo Park and Joseph Rocca and Joshua Johnstun and Joshua Saxe and Junteng Jia and Kalyan Vasuden Alwala and Karthik Prasad and Kartikeya Upasani and Kate Plawiak and Ke Li and Kenneth Heafield and Kevin Stone and Khalid El-Arini and Krithika Iyer and Kshitiz Malik and Kuenley Chiu and Kunal Bhalla and Kushal Lakhotia and Lauren Rantala-Yeary and Laurens van der Maaten and Lawrence Chen and Liang Tan and Liz Jenkins and Louis Martin and Lovish Madaan and Lubo Malo and Lukas Blecher and Lukas Landzaat and Luke de Oliveira and Madeline Muzzi and Mahesh Pasupuleti and Mannat Singh and Manohar Paluri and Marcin Kardas and Maria Tsimpoukelli and Mathew Oldham and Mathieu Rita and Maya Pavlova and Melanie Kambadur and Mike Lewis and Min Si and Mitesh Kumar Singh and Mona Hassan and Naman Goyal and Narjes Torabi and Nikolay Bashlykov and Nikolay Bogoychev and Niladri Chatterji and Ning Zhang and Olivier Duchenne and Onur Çelebi and Patrick Alrassy and Pengchuan Zhang and Pengwei Li and Petar Vasic and Peter Weng and Prajjwal Bhargava and Pratik Dubal and Praveen Krishnan and Punit Singh Koura and Puxin Xu and Qing He and Qingxiao Dong and Ragavan Srinivasan and Raj Ganapathy and Ramon Calderer and Ricardo Silveira Cabral and Robert Stojnic and Roberta Raileanu and Rohan Maheswari and Rohit Girdhar and Rohit Patel and Romain Sauvestre and Ronnie Polidoro and Roshan Sumbaly and Ross Taylor and Ruan Silva and Rui Hou and Rui Wang and Saghar Hosseini and Sahana Chennabasappa and Sanjay Singh and Sean Bell and Seohyun Sonia Kim and Sergey Edunov and Shaoliang Nie and Sharan Narang and Sharath Raparthy and Sheng Shen and Shengye Wan and Shruti Bhosale and Shun Zhang and Simon Vandenhende and Soumya Batra and Spencer Whitman and Sten Sootla and Stephane Collot and Suchin Gururangan and Sydney Borodinsky and Tamar Herman and Tara Fowler and Tarek Sheasha and Thomas Georgiou and Thomas Scialom and Tobias Speckbacher and Todor Mihaylov and Tong Xiao and Ujjwal Karn and Vedanuj Goswami and Vibhor Gupta and Vignesh Ramanathan and Viktor Kerkez and Vincent Gonguet and Virginie Do and Vish Vogeti and Vítor Albiero and Vladan Petrovic and Weiwei Chu and Wenhan Xiong and Wenyin Fu and Whitney Meers and Xavier Martinet and Xiaodong Wang and Xiaofang Wang and Xiaoqing Ellen Tan and Xide Xia and Xinfeng Xie and Xuchao Jia and Xuewei Wang and Yaelle Goldschlag and Yashesh Gaur and Yasmine Babaei and Yi Wen and Yiwen Song and Yuchen Zhang and Yue Li and Yuning Mao and Zacharie Delpierre Coudert and Zheng Yan and Zhengxing Chen and Zoe Papakipos and Aaditya Singh and Aayushi Srivastava and Abha Jain and Adam Kelsey and Adam Shajnfeld and Adithya Gangidi and Adolfo Victoria and Ahuva Goldstand and Ajay Menon and Ajay Sharma and Alex Boesenberg and Alexei Baevski and Allie Feinstein and Amanda Kallet and Amit Sangani and Amos Teo and Anam Yunus and Andrei Lupu and Andres Alvarado and Andrew Caples and Andrew Gu and Andrew Ho and Andrew Poulton and Andrew Ryan and Ankit Ramchandani and Annie Dong and Annie Franco and Anuj Goyal and Aparajita Saraf and Arkabandhu Chowdhury and Ashley Gabriel and Ashwin Bharambe and Assaf Eisenman and Azadeh Yazdan and Beau James and Ben Maurer and Benjamin Leonhardi and Bernie Huang and Beth Loyd and Beto De Paola and Bhargavi Paranjape and Bing Liu and Bo Wu and Boyu Ni and Braden Hancock and Bram Wasti and Brandon Spence and Brani Stojkovic and Brian Gamido and Britt Montalvo and Carl Parker and Carly Burton and Catalina Mejia and Ce Liu and Changhan Wang and Changkyu Kim and Chao Zhou and Chester Hu and Ching-Hsiang Chu and Chris Cai and Chris Tindal and Christoph Feichtenhofer and Cynthia Gao and Damon Civin and Dana Beaty and Daniel Kreymer and Daniel Li and David Adkins and David Xu and Davide Testuggine and Delia David and Devi Parikh and Diana Liskovich and Didem Foss and Dingkang Wang and Duc Le and Dustin Holland and Edward Dowling and Eissa Jamil and Elaine Montgomery and Eleonora Presani and Emily Hahn and Emily Wood and Eric-Tuan Le and Erik Brinkman and Esteban Arcaute and Evan Dunbar and Evan Smothers and Fei Sun and Felix Kreuk and Feng Tian and Filippos Kokkinos and Firat Ozgenel and Francesco Caggioni and Frank Kanayet and Frank Seide and Gabriela Medina Florez and Gabriella Schwarz and Gada Badeer and Georgia Swee and Gil Halpern and Grant Herman and Grigory Sizov and Guangyi and Zhang and Guna Lakshminarayanan and Hakan Inan and Hamid Shojanazeri and Han Zou and Hannah Wang and Hanwen Zha and Haroun Habeeb and Harrison Rudolph and Helen Suk and Henry Aspegren and Hunter Goldman and Hongyuan Zhan and Ibrahim Damlaj and Igor Molybog and Igor Tufanov and Ilias Leontiadis and Irina-Elena Veliche and Itai Gat and Jake Weissman and James Geboski and James Kohli and Janice Lam and Japhet Asher and Jean-Baptiste Gaya and Jeff Marcus and Jeff Tang and Jennifer Chan and Jenny Zhen and Jeremy Reizenstein and Jeremy Teboul and Jessica Zhong and Jian Jin and Jingyi Yang and Joe Cummings and Jon Carvill and Jon Shepard and Jonathan McPhie and Jonathan Torres and Josh Ginsburg and Junjie Wang and Kai Wu and Kam Hou U and Karan Saxena and Kartikay Khandelwal and Katayoun Zand and Kathy Matosich and Kaushik Veeraraghavan and Kelly Michelena and Keqian Li and Kiran Jagadeesh and Kun Huang and Kunal Chawla and Kyle Huang and Lailin Chen and Lakshya Garg and Lavender A and Leandro Silva and Lee Bell and Lei Zhang and Liangpeng Guo and Licheng Yu and Liron Moshkovich and Luca Wehrstedt and Madian Khabsa and Manav Avalani and Manish Bhatt and Martynas Mankus and Matan Hasson and Matthew Lennie and Matthias Reso and Maxim Groshev and Maxim Naumov and Maya Lathi and Meghan Keneally and Miao Liu and Michael L. Seltzer and Michal Valko and Michelle Restrepo and Mihir Patel and Mik Vyatskov and Mikayel Samvelyan and Mike Clark and Mike Macey and Mike Wang and Miquel Jubert Hermoso and Mo Metanat and Mohammad Rastegari and Munish Bansal and Nandhini Santhanam and Natascha Parks and Natasha White and Navyata Bawa and Nayan Singhal and Nick Egebo and Nicolas Usunier and Nikhil Mehta and Nikolay Pavlovich Laptev and Ning Dong and Norman Cheng and Oleg Chernoguz and Olivia Hart and Omkar Salpekar and Ozlem Kalinli and Parkin Kent and Parth Parekh and Paul Saab and Pavan Balaji and Pedro Rittner and Philip Bontrager and Pierre Roux and Piotr Dollar and Polina Zvyagina and Prashant Ratanchandani and Pritish Yuvraj and Qian Liang and Rachad Alao and Rachel Rodriguez and Rafi Ayub and Raghotham Murthy and Raghu Nayani and Rahul Mitra and Rangaprabhu Parthasarathy and Raymond Li and Rebekkah Hogan and Robin Battey and Rocky Wang and Russ Howes and Ruty Rinott and Sachin Mehta and Sachin Siby and Sai Jayesh Bondu and Samyak Datta and Sara Chugh and Sara Hunt and Sargun Dhillon and Sasha Sidorov and Satadru Pan and Saurabh Mahajan and Saurabh Verma and Seiji Yamamoto and Sharadh Ramaswamy and Shaun Lindsay and Shaun Lindsay and Sheng Feng and Shenghao Lin and Shengxin Cindy Zha and Shishir Patil and Shiva Shankar and Shuqiang Zhang and Shuqiang Zhang and Sinong Wang and Sneha Agarwal and Soji Sajuyigbe and Soumith Chintala and Stephanie Max and Stephen Chen and Steve Kehoe and Steve Satterfield and Sudarshan Govindaprasad and Sumit Gupta and Summer Deng and Sungmin Cho and Sunny Virk and Suraj Subramanian and Sy Choudhury and Sydney Goldman and Tal Remez and Tamar Glaser and Tamara Best and Thilo Koehler and Thomas Robinson and Tianhe Li and Tianjun Zhang and Tim Matthews and Timothy Chou and Tzook Shaked and Varun Vontimitta and Victoria Ajayi and Victoria Montanez and Vijai Mohan and Vinay Satish Kumar and Vishal Mangla and Vlad Ionescu and Vlad Poenaru and Vlad Tiberiu Mihailescu and Vladimir Ivanov and Wei Li and Wenchen Wang and Wenwen Jiang and Wes Bouaziz and Will Constable and Xiaocheng Tang and Xiaojian Wu and Xiaolan Wang and Xilun Wu and Xinbo Gao and Yaniv Kleinman and Yanjun Chen and Ye Hu and Ye Jia and Ye Qi and Yenda Li and Yilin Zhang and Ying Zhang and Yossi Adi and Youngjin Nam and Yu and Wang and Yu Zhao and Yuchen Hao and Yundi Qian and Yunlu Li and Yuzi He and Zach Rait and Zachary DeVito and Zef Rosnbrick and Zhaoduo Wen and Zhenyu Yang and Zhiwei Zhao and Zhiyu Ma},
      year={2024},
      eprint={2407.21783},
      archivePrefix={arXiv},
      primaryClass={cs.AI},
      url={https://arxiv.org/abs/2407.21783}, 
}

@inproceedings{
hu2022lora,
title={Lo{RA}: Low-Rank Adaptation of Large Language Models},
author={Edward J Hu and yelong shen and Phillip Wallis and Zeyuan Allen-Zhu and Yuanzhi Li and Shean Wang and Lu Wang and Weizhu Chen},
booktitle={International Conference on Learning Representations},
year={2022},
url={https://openreview.net/forum?id=nZeVKeeFYf9}
}

@inproceedings{10.5555/3666122.3668142,
author = {Zheng, Lianmin and Chiang, Wei-Lin and Sheng, Ying and Zhuang, Siyuan and Wu, Zhanghao and Zhuang, Yonghao and Lin, Zi and Li, Zhuohan and Li, Dacheng and Xing, Eric P. and Zhang, Hao and Gonzalez, Joseph E. and Stoica, Ion},
title = {Judging LLM-as-a-judge with MT-bench and Chatbot Arena},
year = {2023},
publisher = {Curran Associates Inc.},
address = {Red Hook, NY, USA},
booktitle = {Proceedings of the 37th International Conference on Neural Information Processing Systems},
articleno = {2020},
numpages = {29},
location = {New Orleans, LA, USA},
series = {NIPS '23}
}

@inproceedings{
hsieh2024ruler,
title={{RULER}: What{\textquoteright}s the Real Context Size of Your Long-Context Language Models?},
author={Cheng-Ping Hsieh and Simeng Sun and Samuel Kriman and Shantanu Acharya and Dima Rekesh and Fei Jia and Boris Ginsburg},
booktitle={First Conference on Language Modeling},
year={2024},
url={https://openreview.net/forum?id=kIoBbc76Sy}
}

@inproceedings{lin-2004-rouge,
    title = "{ROUGE}: A Package for Automatic Evaluation of Summaries",
    author = "Lin, Chin-Yew",
    booktitle = "Text Summarization Branches Out",
    month = jul,
    year = "2004",
    address = "Barcelona, Spain",
    publisher = "Association for Computational Linguistics",
    url = "https://aclanthology.org/W04-1013/",
    pages = "74--81"
}

@inproceedings{
topping2022understanding,
title={Understanding over-squashing and bottlenecks on graphs via curvature},
author={Jake Topping and Francesco Di Giovanni and Benjamin Paul Chamberlain and Xiaowen Dong and Michael M. Bronstein},
booktitle={International Conference on Learning Representations},
year={2022},
url={https://openreview.net/forum?id=7UmjRGzp-A}
}

@inproceedings{
alon2021on,
title={On the Bottleneck of Graph Neural Networks and its Practical Implications},
author={Uri Alon and Eran Yahav},
booktitle={International Conference on Learning Representations},
year={2021},
url={https://openreview.net/forum?id=i80OPhOCVH2}
}

@inproceedings{
Zhang*2020BERTScore:,
title={BERTScore: Evaluating Text Generation with BERT},
author={Tianyi Zhang* and Varsha Kishore* and Felix Wu* and Kilian Q. Weinberger and Yoav Artzi},
booktitle={International Conference on Learning Representations},
year={2020},
url={https://openreview.net/forum?id=SkeHuCVFDr}
}

\appendix
\section{Graph Construction}
\label{app:graph-constr}
Formally, each paper is represented as an undirected graph $\mathcal{G} = (\mathcal{V}, \mathcal{E})$, where nodes $\mathcal{V}$ represent text segments with semantically encoded features, and $\mathcal{E}$ denote the hierarchical and sequential relationships that form the document's structure. Our graph construction is designed to be generalizable across diverse academic paper structures, accommodating arbitrary levels of nesting and heterogeneous section organization.

\subsection{Node Definitions}
\begin{itemize}
  \item \textbf{Paper Node} ($v_{\text{paper}}$): A single root node representing the entire document.
  \item \textbf{Heading Nodes} ($\mathcal{H} = \{h_1, h_2, \dots, h_m\}$): Nodes corresponding to top-level section headings.
  \item \textbf{Sub-heading Nodes} ($\mathcal{S}_i = \{s_{i1}, s_{i2}, \dots, s_{ik}\}$): Nodes representing sub-headings under each heading $h_i$. This structure naturally extends to accommodate multiple levels of nesting, making it suitable for documents with deeper hierarchical organization.
  \item \textbf{Passage Nodes} ($\mathcal{O}_j = \{o_{j1}, o_{j2}, \dots, o_{jn}\}$): Nodes representing text segments, split by newline characters (\textbackslash n), capturing logical textual blocks.
  \item \textbf{Sentence Nodes} ($\mathcal{T}_{jk} = \{t_{jk1}, t_{jk2}, \dots, t_{jkp}\}$): Nodes for individual sentences within each passage $o_{jk}$, tokenized using NLTK \cite{bird-loper-2004-nltk}.
\end{itemize}

\subsection{Edge Definitions}
Edges capture both hierarchical and sequential relationships within the document.

\begin{itemize}
  \item \textbf{Root Connections}: $(v_{\text{paper}}, h_i) \in \mathcal{E}_{\text{hier}}$ for all $h_i \in \mathcal{H}$.
  \item \textbf{Heading-SubHeading Connections}:
  
  $(h_i, s_{ij}) \in \mathcal{E}_{\text{hier}}$ for all $s_{ij} \in \mathcal{S}_i$.
  \item \textbf{Content Links}: 
  
  $(s_{ij}, o_{jk}) \in \mathcal{E}_{\text{hier}} \text{ if } \mathcal{S}_i \neq \emptyset; (h_i, o_{jk}) \in \mathcal{E}_{\text{hier}} \text{ otherwise.}$
  \item \textbf{Passage-Sentence Connections}: 
  
  $(o_{jk}, t_{jkq}) \in \mathcal{E}_{\text{hier}}$ for all $t_{jkq} \in \mathcal{T}_{jk}$.
  \item \textbf{Sequential Edges}:
  
    For passages: $(o_{jk}, o_{j(k+1)}) \in \mathcal{E}_{\text{seq}}$ for $1 \leq k < n$.  
    
    For sentences: $(t_{jkq}, t_{jk(q+1)}) \in \mathcal{E}_{\text{seq}}$ for $1 \leq q < p$.
\end{itemize}

\noindent The final edge set is the union of hierarchical and sequential edges:
\begin{equation}
  \mathcal{E} = \mathcal{E}_{\text{hier}} \cup \mathcal{E}_{\text{seq}},
\end{equation}
where $\mathcal{E}_{\text{hier}}$ captures the vertical (structural) relationships and $\mathcal{E}_{\text{seq}}$ captures the horizontal (textual progression) flow. 
This graph construction is agnostic to section depth and flexibly adapts to any nested structure within academic documents. A toy example of such a graph, built from a paper with 2 Headings, 2 Sub-Headings, 4 Passages, and 6 Sentences, is shown in Figure~\ref{fig:toy-example}.
\section{Ablation studies}
\label{sec:abl-studies}
The quantitative performance of different combinations of \(k\) (number of review sentences concatenated; \(k = 1, 3, 5\)) and \(m\) (number of top DPR-retrieved passages used to train the GNN; \(m = 3, 5\)) for review generation is reported in Table~\ref{tab:mistral-ablation} and Table~\ref{tab:llama-ablation}, evaluated against the reference feedback. While calculating the ROUGE metrics, we enable the \texttt{use\_stemmer} option to ensure consistent matching of words with different morphological forms, to improve the reliability of the evaluation.
\begin{table*}[htbp]
\centering
\begin{tabular}{l ccc ccc c c}
\toprule
Model 
& \multicolumn{3}{c}{ROUGE-F1} 
& \multicolumn{3}{c}{ROUGE-Recall} 
& BERTScore F1 & Training Time (hrs) \\
\cmidrule(lr){2-4} \cmidrule(lr){5-7}
& R-1 & R-2 & R-L 
& R-1 & R-2 & R-L 
& & \\
\midrule
Mistral \((1, 3)\) & 53.54 & 15.81 & \underline{22.68} & 53.26 & 15.74 & \underline{22.59} & 86.51 & 7.60 \\
Mistral \((1, 5)\) & 53.54 & 15.76 & 22.64 & 52.81 & 15.56 & 22.36 & 86.51 & 10.22 \\
Mistral \((3, 3)\) & \underline{53.86} & 15.80 & 22.62 & 53.21 & 15.63 & 22.36 & 86.54 & 6.33 \\
Mistral \((3, 5)\) & \textbf{53.87} & \textbf{15.91} & 22.74 & \underline{53.43} & \underline{15.79} & 22.57 & \underline{86.54} & 7.64 \\
Mistral \((5, 3)\) & 53.44 & 15.47 & 22.39 & 53.18 & 15.42 & 22.31 & 86.44 & 4.34 \\
Mistral \((5, 5)\) & 53.65 & \underline{15.80} & \textbf{22.74} & \textbf{53.60} & \textbf{15.79} & \textbf{22.74} & \textbf{86.55} & 6.44 \\
\bottomrule
\end{tabular}
\caption{Overall performance of Mistral across different $(k, m)$ settings. Metrics include ROUGE-F1 (R-1, R-2, R-L), ROUGE-Recall (R-1, R-2, R-L), BERTScore F1 and training time (in hours). Bold highlights the best scores, and underlined values indicate the second-best.}
\label{tab:mistral-ablation}
\end{table*}

\begin{table*}[htbp]
\centering
\begin{tabular}{l ccc ccc c c}
\toprule
Model 
& \multicolumn{3}{c}{ROUGE-F1} 
& \multicolumn{3}{c}{ROUGE-Recall} 
& BERTScore F1 & Training Time (hrs) \\
\cmidrule(lr){2-4} \cmidrule(lr){5-7}
& R-1 & R-2 & R-L 
& R-1 & R-2 & R-L 
& & \\
\midrule
Llama \((1, 3)\) & 54.42 & 16.19 & 22.76 & 53.71 & 15.98 & 22.48 & 86.61 & 6.65 \\
Llama \((1, 5)\) & 54.40 & 16.09 & 22.82 & 53.58 & 15.86 & \underline{22.50} & 86.61 & 8.23 \\
Llama \((3, 3)\) & \underline{54.52} & 16.03 & 22.67 & \textbf{53.79} & 15.83 & 22.39 & 86.61 & 5.11 \\
Llama \((3, 5)\) & \textbf{54.61} & \textbf{16.30} & \textbf{22.88} & \underline{53.78} & \textbf{16.06} & \textbf{22.56} & \underline{86.66} & 6.28 \\
Llama \((5, 3)\) & 54.24 & 15.91 & 22.59 & 53.75 & 15.77 & 22.41 & 86.59 & 3.60 \\
Llama \((5, 5)\) & 54.47 & \underline{16.26} & \underline{22.87} & 53.49 & \underline{15.98} & 22.48 & \textbf{86.68} & 5.30 \\
\bottomrule
\end{tabular}
\caption{Overall performance of Llama across different $(k, m)$ settings. Metrics include ROUGE-F1 (R-1, R-2, R-L), ROUGE-Recall (R-1, R-2, R-L), BERTScore F1 and training time (in hours). Bold highlights the best scores, and underlined values indicate the second-best.}
\label{tab:llama-ablation}
\end{table*}
\section{GNN Implementation Details}
\label{sec:appendix-gnn}
The passing of information from one node to another is enhanced by the GATConv (Equations~\ref{eq:GATConv}, \ref{eq:GATConv_attention_1}, and \ref{eq:GATConv_attention_2}) layers, which involves the concept of attention to selectively propagate messages in the Message Passing layer of a GNN.
\begin{equation}
\mathbf{h}_i' = \sigma \left( \sum_{j \in \mathcal{N}(i)} 
\alpha_{ij} \mathbf{W} \mathbf{h}_j \right)
\label{eq:GATConv}
\end{equation}

\begin{equation}
e_{ij} = \text{LeakyReLU} \Bigl( \mathbf{a}^T \bigl[\mathbf{W} \mathbf{h}_i \parallel \mathbf{W} \mathbf{h}_j \bigr] \Bigr)
\label{eq:GATConv_attention_1}
\end{equation}

\begin{equation}
\alpha_{ij} = \frac{\exp\bigl(e_{ij}\bigr)}{\sum_{k \in \mathcal{N}(i)} \exp\bigl(e_{ik}\bigr)}
\label{eq:GATConv_attention_2}
\end{equation}
We train and test our model on a single NVIDIA RTX6000 GPU with 48GB of VRAM.
\subsection{Model Architecture}
The model contains a Graph Neural Network (GNN) module and a passage classifier.
\paragraph{GNN Module}
The GNN uses three graph attention (\texttt{GATConv}) layers. The first \texttt{GATConv} layer processes input features (of size \textit{768 (embedding dimension of Sentence Encoder)}) into 128-dimensional features with 4 attention heads, producing a concatenated output of $128 \times 4$ dimensions. A residual connection is incorporated by applying a linear layer to project the input to $128 \times 4$, which is added to the \texttt{GATConv} output. This is followed by an ELU activation and dropout ($p=0.2$).

The second \texttt{GATConv} layer maps the $128 \times 4$ input to 256 dimensional features using 2 attention heads (output dimension $256 \times 2$). A residual connection similarly projects the prior layer's output to $256 \times 2$, followed by ELU and dropout ($p=0.2$). The final \texttt{GATConv} layer upscales the $256 \times 2$ features to the target output dimension (\texttt{1024 in our case}) without residual connections and dropout layers.

\paragraph{Passage Classifier}
For passage nodes, we employ a feedforward classifier to map the output embeddings from the GNN to the final label space. The classifier is implemented as a multi-layer perceptron (MLP) comprising the following sequence of operations:

\begin{itemize}
\item A linear transformation projecting the input to a 512 dimensional space.
\item A SiLU activation followed by dropout with a probability of $p=0.3$.
\item A second linear layer projecting to 256 dimensions.
\item Layer normalization applied to the 256-dimensional features.
\item Another SiLU activation.
\item A final linear layer projecting to the number of target classes (typically 2 for binary classification).
\end{itemize}

This architecture enables the model to learn nonlinear transformations of the graph-encoded passage representations while incorporating regularization and normalization for stable training.

\subsection{Training}
The model is trained for 30 epochs, with the best validation performance checkpoint retained for inference. The loss function used for training is the Cross Entropy Loss.

\section{Finetuning Implementation Details}
\label{sec:appendix-ft}
We fine-tune \textit{Llama-3.1-8B} using two NVIDIA L40S GPUs and \textit{Mistral} using a single NVIDIA RTX 6000 Ada GPU. Both GPUs are equipped with 48GB of GDDR6 VRAM.

We apply Low-Rank Adaptation (LoRA) with a rank of 8 and a scaling factor ($\alpha = 16$). LoRA is integrated into the following transformer layers: \textit{q\_proj}, \textit{k\_proj}, \textit{v\_proj}, \textit{o\_proj}, \textit{gate\_proj}, \textit{up\_proj}, and \textit{down\_proj}. Models are fine-tuned for 3 epochs. The total training time for different configurations is reported in Table~\ref{tab:mistral-ablation} and Table~\ref{tab:llama-ablation}. The full set of hyperparameters used during training is listed in Table~\ref{tab:hyperparams}.

\begin{table}[htbp]
\centering
\begin{tabular}{ll}
\toprule
\textbf{Hyperparameter} & \textbf{Value} \\
\midrule
Batch Size & 1 \\
Gradient Accumulation Steps & 1 \\
16-bit Floating Point Precision & True \\
Optimizer & adamw\_torch \\
Learning Rate & $1 \times 10^{-4}$ \\
Max Gradient Norm & 0.3 \\
Warmup Ratio & 0.03 \\
LoRA Rank & 8 \\
LoRA scaling factor & 16 \\
LoRA Dropout & 0.05 \\
Weight Decay & 0.01 \\
LR Scheduler Type & cosine \\
\bottomrule
\end{tabular}
\caption{Hyperparameters Used During Finetuning}
\label{tab:hyperparams}
\end{table}

\subsection{Prompt for Finetuning}
\label{sec:appendix-prompt-ft}
The prompt used for fine-tuning both \textit{Llama 3.1 8B} and \textit{Mistral-Instruct-v0.2} is shown in Listing~\ref{lst:prompt}.

\begin{tcolorbox}[title={Prompt Used During Finetuning}, width=\linewidth, colback=white, colframe=gray, arc=0pt, outer arc=5pt, boxrule=0.5pt, leftrule=2pt, rightrule=2pt, right=0pt, left=0pt, top=0pt, bottom=0pt, toprule=0pt, bottomrule=2pt]
\label{lst:prompt}
\small
Below is an instruction that describes a task,
paired with an input that provides further
context. Write a response that appropriately
completes the request.

\textbf{\#\#\# Instruction:}

Generate a structured feedback for the research
paper passages provided below. The feedback
should include a summary of the paper, its
strengths, weaknesses, and questions for
the authors. Consider that the feedback is being
given for a paper submitted to the ICLR conference.

\textbf{\#\#\# Research Paper Passages:}

\textbf{\#\#\# Feedback for the paper:}

\texttt{**Summary** + "\textbackslash n\textbackslash n" +}

\texttt{**Strengths** + "\textbackslash n\textbackslash n" +}

\texttt{**Weaknesses** + "\textbackslash n\textbackslash n" +}

\texttt{**Questions** + "\textbackslash n\textbackslash n"}
\end{tcolorbox}
The extracted relevant passages are inserted ahead of the "Research Paper Passages" heading. Similarly, the corresponding content for each feedback section (Summary, Strengths, Weaknesses, and Questions) is added after their respective headings in the generated feedback.

\section{LLM-as-a-judge and Qualitative Assessment}
\label{app:llm-as-a-judge}
The prompt used for LLM-as-a-judge with Gemini-2.0-Flash is shown in Figure~\ref{llmjudge:prompt}. The corresponding model-generated feedback is inserted under the appropriate subheadings within the prompt.

For Qualitative Assessment, the prompt in Figure~\ref{llmjudge:prompt} is modified only slightly as to just remove the output format. This slightly modified prompt is given as the evaluation guideline to the annotators. It is ensured that the annotators selected are eligible to review papers for the particular conference.
\begin{figure*}[htbp]
\small
\begin{verbatim}
** Instruction ** : You are an expert on research papers. Evaluate the following feedback according to the scoring 
methodology. \n Given below is a structured feedback generated for a paper submitted to the ICLR conference. \n\n
Give rating for the categories described below on the full scale from 1(lowest rating) to 5(highest rating). 
Assign scores independently per category (they may differ significantly).

**Category 1: Feedback Depth**
1 – Unqualified: Admits lack of expertise | Fundamental misunderstandings | Fails to engage with technical content
2 – Limited: Surface-level comprehension | Misses >2 key concepts | Confuses established methods
3 – Functional: Grasps main contributions | Identifies 1-2 technical flaws | Limited literature context
4 – Qualified: Precise technical critique | Contextualizes in recent work (>3 citations) | 
Identifies methodological tradeoffs
5 – Authoritative: Demonstrates field leadership | Predicts follow-up research directions | 
Reveals fundamental limitations in approach

**Category 2: Feedback Thoroughness**
1 – Negligent: <3 substantive comments | Skips methodology/results | Fails to address claims
2 – Patchy: Only addresses obvious aspects | 0 analysis of experiments | No ethical consideration
3 – Standard: Checks main claims | Basic methodology feedback | 1 reproducibility concern noted
4 – Rigorous: Line-by-line methodology check | Statistical validity analysis | 2+ improvement suggestions
5 – Forensic: Alternative approach proposals | Re-analysis of key results | 
Compliance checklist (ethics, reproducibility, etc.)

**Category 3: Feedback Constructiveness**
1 – Destructive: Personal attacks | Unsubstantiated dismissal | 0 actionable feedback
2 – Demanding: Unrealistic requests ("redo all experiments") | Vague quality complaints
3 – Directive: General improvement areas | 1-2 specific examples | Mixed positive/negative tone
4 – Strategic: Priority-ranked suggestions | Alternative methodology paths | Conference-specific improvements
5 – Transformative: Step-by-step revision roadmap | Code/experiment snippets provided | Before-after examples

**Category 4: Feedback Helpfulness**
1 – Misleading: Factually incorrect suggestions | Contradicts paper's goals | Harmful recommendations
2 – Abstract: "Improve writing" without examples | "Add experiments" without design 
3 – Guided: Section-specific edits | 2-3 implementable suggestions | Partial error diagnosis
4 – Impactful: Directly address weaknesses | Templates/examples provided | Clear priority levels
5 – Revelatory: Exposes fundamental flaws | Provides benchmark comparisons | Supplies missing literature

**Scoring Methodology**
For EACH category:
1. Find the HIGHEST level (5-1) where ALL criteria are FULLY satisfied (Base Level)
2. Check criteria in the NEXT HIGHER LEVEL (Base Level + 1):
   - Calculate partial credit: (Satisfied Criteria in Higher Level / Total Criteria in Higher Level)
3. Final Score = Base Level + Partial Credit
4. Round to 2 decimal places (0.33 increments for 3-criteria levels, 0.5 for 2-criteria levels)

** Structured Feedback ** : {structured_feedback} \n\n

** Scores for the paper feedback ** :\n
Respond ONLY with this JSON format:
{{
  "feedback_scores": {{
    "Category 1": "",
    "Category 2": "",
    "Category 3": "",
    "Category 4": ""
  }}
}}
\end{verbatim}
\caption{Prompt used during evaluating the quality of generated feedbacks with LLM as a Judge}
\label{llmjudge:prompt}
\end{figure*}
\section{Human Evaluation UI}
\label{app:human_eval_ui}
Our annotation UI is developed using Streamlit \footnote{https://streamlit.io/}.

\subsection{Qualitative Assessment}
\begin{figure*}[htbp]
  \centering
  \includegraphics[width=0.85\linewidth]{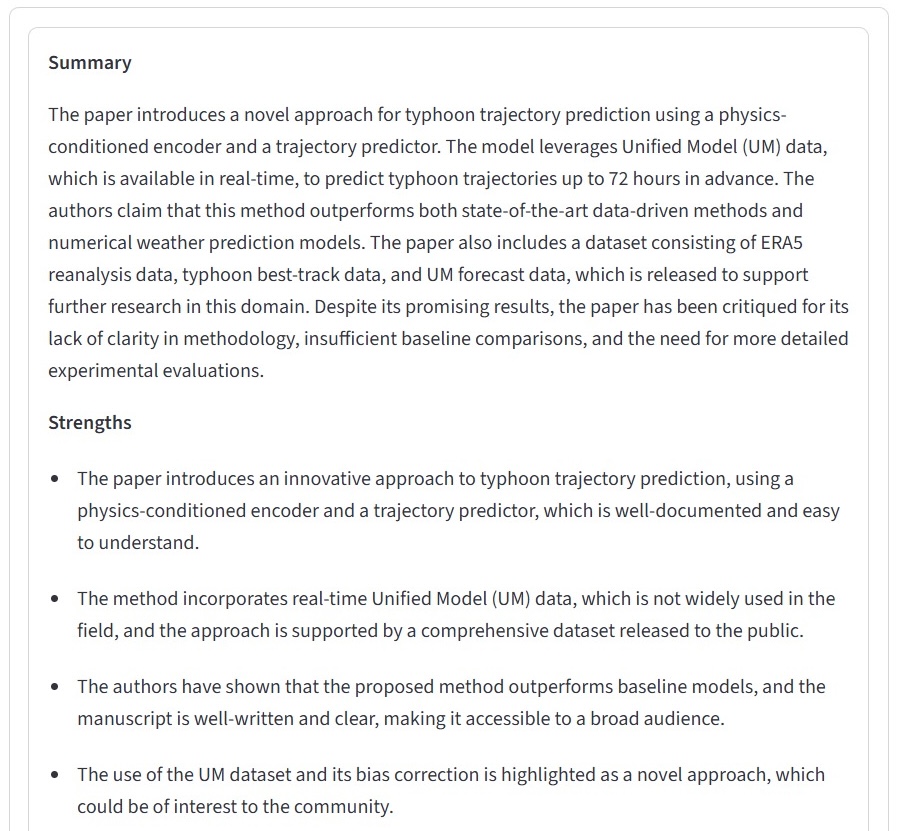}
  \caption{This interface component corresponds to the Qualitative Assessment task described in Section~\ref{sec:human-qualitative}. It marks the beginning of the feedback presentation. Due to space constraints, only the summary and a subset of the strengths from one feedback are shown here.}
  \label{fig:HE_1_1}
\end{figure*}

\begin{figure*}[htbp]
  \centering
  \includegraphics[width=0.885\linewidth]{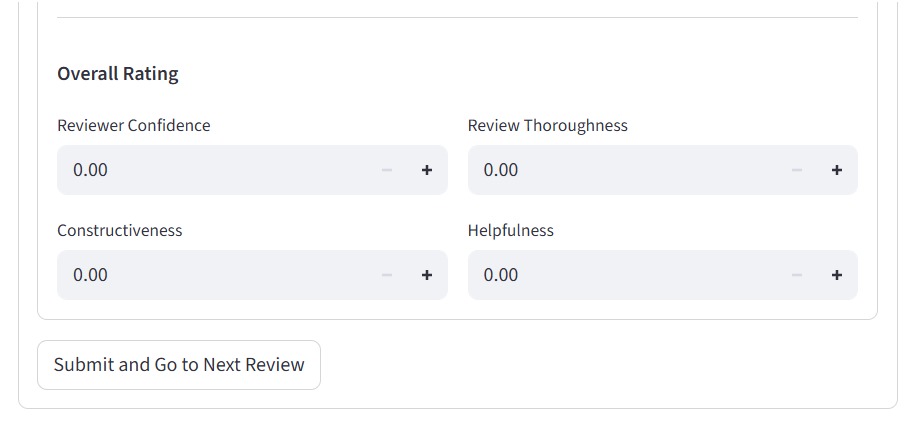}
  \caption{This interface component corresponds to the task described in Section~\ref{sec:human-qualitative}. In this part, annotators provide a final rating across the four categories and submit their annotation.}
  \label{fig:HE_1_2}
\end{figure*}

Examples of the user interface used for the Qualitative Assessment are shown in Figure~\ref{fig:HE_1_1} and ~\ref{fig:HE_1_2}.

\label{app:qualitative_assessment}

\subsection{Factual Validity Assessment}
\begin{figure*}
\centering
\subfloat[The original paper is displayed in a PDF viewer to provide factual context.]{\includegraphics[width=0.3825\textwidth]{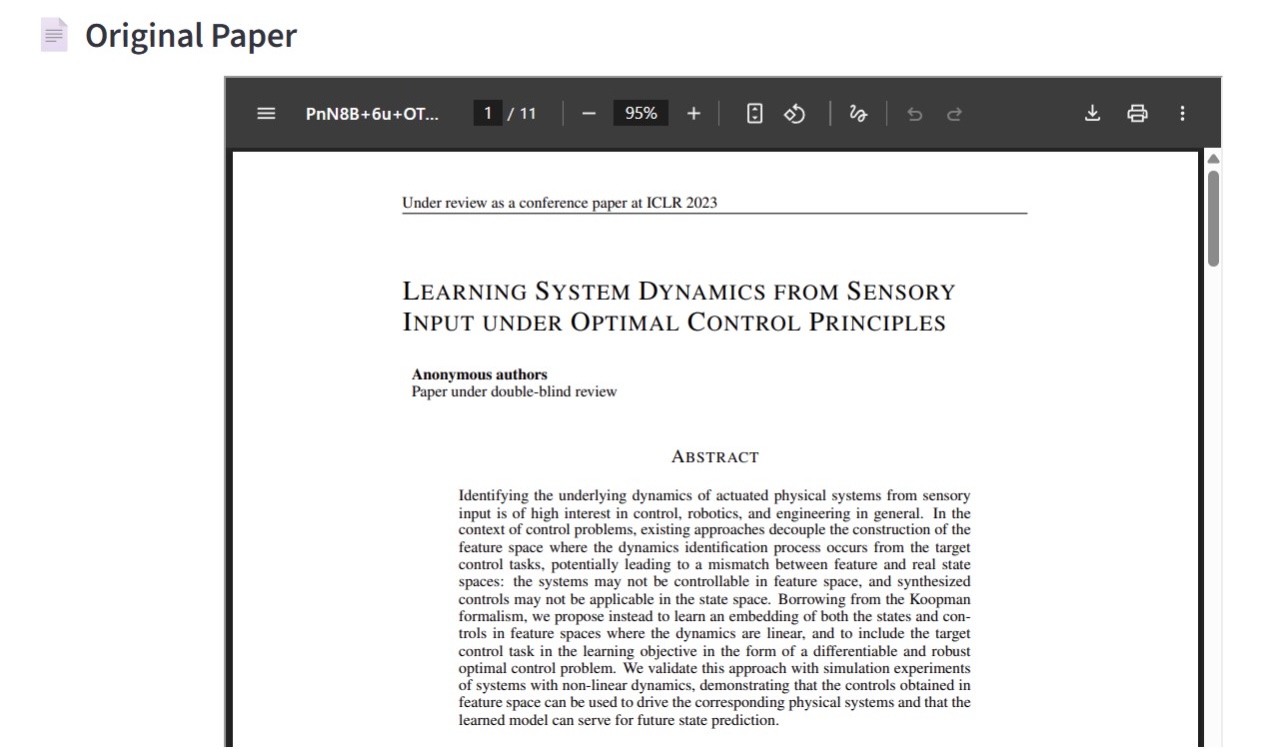}\label{fig:HE_2_1}}
\hfill
\subfloat[Each point of the review is rated for accuracy using a dropdown menu.]{\includegraphics[width=0.3825\textwidth]{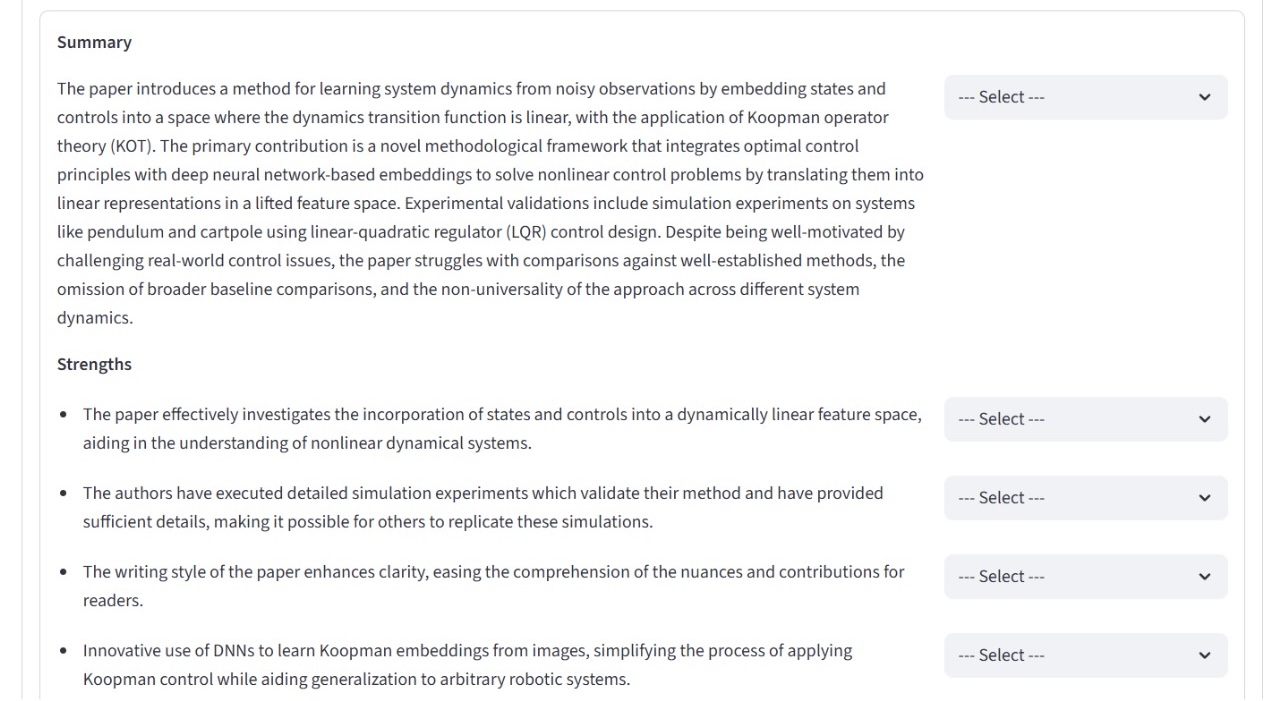}\label{fig:HE_2_2}}
\caption{The user interface for the Factual Validity Assessment task described in Section~\ref{sec:factual}. On the left (a), the full research paper is presented to the annotator. On the right (b), the annotator rates each point of the feedback based on its consistency with the source document.}
\label{fig:factual_assessment_ui}
\end{figure*}

\begin{table}[htbp]
  \centering
  \setlength{\tabcolsep}{3.5pt}
  \begin{tabular}{llc}
    \toprule
    \textbf{Dataset} & \textbf{Feedback} & \textbf{Positive Rating (\%)} \\
    \midrule
    \multirow{3}{*}{\shortstack[l]{ICLR\\2024}} 
      & ICT\_10 & 76.97 \\
      & Llama \((5, 3)\) & 65.25 \\
      & Llama \((5, 5)\) & 64.58 \\
    \midrule
    \multirow{2}{*}{Other} 
      & Llama \((5, 3)\) & 61.86 \\
      & Llama \((5, 5)\) & 64.31 \\
    \bottomrule
  \end{tabular}
  \caption{Overall Positive Rating Percentage (sum of ``Completely Agree'' and ``Mostly Agree'') for test dataset of ICLR 2024 as well as NeurIPS and COLM 2025 (Other).}
  \label{tab:pos_rating}
\end{table}
Examples of the user interface used for Factual Validity Assessment can be seen in Figures~\ref{fig:HE_2_1} and ~\ref{fig:HE_2_2}.

\section{Evaluation Guidelines for Factual Validity Assessment}
\label{app:eval_guidelines_human}
The evaluation guidelines for the factual validity assessment are shown in Figure~\ref{fig:eval_guideline_fva}. It is ensured that the annotators selected are eligible to review papers for the particular conference.
\begin{figure*}[htbp]
\small
\begin{verbatim}
**Instructions**:
To assess a structured feedback and check how accurately it reflects the paper's content.

**Evaluation Process**:
1. Read the entire paper first.
   - If you need to download the paper for some reason, do ensure that you only download the paper using the provided UI.
2. Read the structured feedback and identify every DISCRETE CLAIM.
   - For the “Summary” section of the review, treat the entire content of the Summary
     as ONE discrete claim.
   - For the “Strengths”, “Weaknesses”, and “Questions” sections of the review,
     treat EACH bulleted point as a separate discrete claim.
3. For each claim, choose the best category from the dropdown.

**Claim Categorization**:
You must rate the Validity Agreement of every claim on a scale of 1–4,
where 1 indicates “Completely Disagree” and 4 indicates “Completely Agree”.

**Completely Disagree (Invalid Claim) [1]**:
- The claim is factually wrong, contradicts the paper, or is hallucinated.

**Mostly Disagree (Largely Invalid Claim) [2]**:
- The claim is mostly incorrect, though it may contain minor truth.

**Mostly Agree (Largely Valid Claim) [3]**:
- The claim is mostly correct, with minor gaps or oversights.

**Completely Agree (Fully Valid Claim) [4]**:
- The claim is entirely accurate.
\end{verbatim}
\caption{Guidelines used for human evaluation of factual validity of the structured feedback}
\label{fig:eval_guideline_fva}
\end{figure*}
\section{Example of a Generated Structured Feedback}
\label{sec:appendix-example}
We present a structured feedback generated by our system in Figure~\ref{fig:feedback-listing}
\begin{figure*}[t!]
    \begin{minipage}{\linewidth}
    
    \small
    \begin{verbatim}
On the Role of Edge Dependency in Graph Generative Models

Summary
The paper introduces a novel framework for evaluating generative models of graphs,
focusing on the concept of ``overlap'' to ensure both accuracy and edge diversity.
It categorizes graph generative models into three levels of complexity: edge
independent, node independent, and fully dependent models. The paper derives
theoretical bounds on triangle counts and other short-length cycles based on overlap
and provides new generative models for each level. However, the paper's novelty and
practical contributions are questioned due to the use of existing methods and the
lack of significant empirical validation. The experimental section is seen as
limited and does not convincingly demonstrate the superiority of the proposed
methods over existing ones.

Strengths
- The paper introduces a novel evaluation framework that considers the importance of
  model-generated graph overlap to ensure both accuracy and edge diversity, which is
  a significant contribution to the field of graph generative models.
- The paper presents a hierarchy of graph generative models categorized into three
  levels of complexity, which encapsulates a wide range of prevalent methods and
  provides a structured framework for evaluating graph generative models.
- The theoretical bounds on the number of triangles and other short-length cycles are
  well-presented and support the findings with instances demonstrating the asymptotic
  optimality of these bounds.
- The authors introduce new generative models for each of the three hierarchical
  levels, leveraging dense subgraph discovery (Gionis & Tsourakakis, 2015), which
  is a practical application of the theoretical findings.

Weaknesses
- The novelty of the paper is limited as it primarily relies on existing methods and
  theoretical frameworks, and the experimental results do not convincingly
  demonstrate the superiority of the proposed methods over existing ones.
- The empirical evaluation is weak and does not provide sufficient evidence to
  support the claims made in the paper. The experimental results are not robust and
  the paper does not provide enough comparisons with other models to establish the
  effectiveness of the proposed methods.
- The paper has several presentation issues, including unclear definitions and
  notations, which could hinder the understanding of the content for readers not
  familiar with graph generative models.
- The practical applications and the significance of the proposed methods are not
  clearly demonstrated, and the paper lacks a detailed discussion on the limitations
  and potential future work.

Questions
- Can the authors clarify the definitions and notations used in the paper,
  particularly the ones related to the overlap and triangle density, to ensure a
  clearer understanding of the content?
- How do the authors justify the novelty of their work given the reliance on
  existing methods and theoretical frameworks?
- What specific contributions does the paper make to the field of graph generative
  models, and can these be clearly articulated?
- Why was the empirical evaluation conducted in a limited manner, and how can the
  paper be improved to provide more robust and convincing experimental results?
- Could the authors discuss the practical applications of the proposed methods and
  the significance of their work in more detail?
- Are there any plans to extend the theoretical bounds to higher-order cycles, and
  how might this affect the overall conclusions drawn in the paper?
    \end{verbatim}
    
    \caption[Output of the generated feedback by AutoRev for the paper On the Role of Edge Dependency in Graph Generative Models]{Output of the generated feedback by AutoRev for the paper \textit{On the Role of Edge Dependency in Graph Generative Models}\footnote{https://openreview.net/forum?id=LCQ7YTzgRQ}}
    \label{fig:feedback-listing}
    
    \end{minipage}
\end{figure*}

\end{document}